\documentclass[runningheads]{llncs}
\usepackage[utf8]{inputenc}
\usepackage{amssymb}
\usepackage{amsmath}
\usepackage{epsfig}
\usepackage{graphicx}
\usepackage{paralist}
\usepackage{amsmath}
\usepackage{amssymb}
\usepackage{amsfonts}
\usepackage[dvipsnames]{xcolor}
\usepackage{colortbl}
\usepackage[percent]{overpic}
\usepackage{microtype}
\usepackage{arydshln}
\usepackage{soul}
\usepackage{float}
\usepackage{balance}
\usepackage{tikz}
\usepackage[utf8]{inputenc}
\usepackage{pgfplots}
\usepackage{color}
\usepackage{lgreek}
\usepackage[hidelinks]{hyperref}
\usepackage[normalem]{ulem}

\DeclareUnicodeCharacter{2212}{−}
\usepgfplotslibrary{groupplots,dateplot}
\usetikzlibrary{patterns,shapes.arrows}
\pgfplotsset{compat=newest}
\usepackage{cuted}
\usepackage{capt-of}
\usepackage{subcaption}
\usepackage{wrapfig,lipsum,booktabs}
\usepackage{etoolbox}
\usepackage[accsupp]{axessibility}
\usepackage{pifont}
\usepackage{multirow}
\usepackage{minibox}
\makeatother

\newcommand{\straighttheta}{\text{\begin{greek}\textbf{j}\end{greek}}}
\newcommand{\x}{{\textbf{x}}} 
\newcommand{\z}{{\textbf{z}}} 
\newcommand{\V}{{\textbf{V}}} 
\newcommand{\etal}{{\textit{et al.}}}
\newcommand{\ie}{{\textit{i.e.}}}
\newcommand{\eg}{{\textit{e.g.}}}
\newcommand{\n}{{\textbf{n}}} 
 
\newcommand{\na}{--}
\newcommand{\best}{\cellcolor{orange!15}}
\newcommand{\worst}{\cellcolor{red!10}}
\newcommand{\pphi}{\Phi}
\newcommand{\numcircle}[1]{\raisebox{.5pt}{\textcircled{\raisebox{-.9pt}{#1}}}}

\definecolor{customdeepgreen}{HTML}{006A71}
\newtheorem{result}{Result}

\makeatletter
  \newcommand\reduline{\bgroup\markoverwith{\textcolor{red}{\rule[-0.5ex]{2pt}{0.4pt}}}\ULon}
  \UL@protected\def\blueuwave{\leavevmode \bgroup 
    \ifdim \ULdepth=\maxdimen \ULdepth 3.5\p@
    \else \advance\ULdepth2\p@ 
    \fi \markoverwith{\lower\ULdepth\hbox{\textcolor{blue}{\sixly \char58}}}\ULon}
  \UL@protected\def\yellowdotuline{\leavevmode \bgroup 
    \UL@setULdepth
    \ifx\UL@on\UL@onin \advance\ULdepth2\p@\fi
    \markoverwith{\begingroup
       \lower\ULdepth\hbox{\kern.06em \textcolor{yellow}{.}\kern.04em}%
       \endgroup}%
    \ULon}
  \UL@protected\def\greendashuline{\leavevmode \bgroup 
    \UL@setULdepth
    \ifx\UL@on\UL@onin \advance\ULdepth2\p@\fi
    \markoverwith{\kern.13em
    \vtop{\color{customdeepgreen}\kern\ULdepth \hrule width .3em}%
    \kern.13em}\ULon}
\makeatother

\begin{document}
\pagestyle{headings}
\mainmatter
\title{A Level Set Theory for Neural Implicit Evolution under
Explicit Flows}
\titlerunning{A Level Set Theory for Neural Implicit Evolution under
Explicit Flows}
\author{Ishit Mehta \qquad Manmohan Chandraker \qquad Ravi Ramamoorthi}
\authorrunning{I. Mehta et al.}
\institute{University of California San Diego}

\maketitle
\begin{abstract}
Coordinate-based neural networks parameterizing implicit surfaces have emerged
as efficient representations of geometry.  They effectively act as parametric
level sets with the zero-level set defining the surface of interest. We present
a framework that allows applying deformation operations defined for
triangle meshes onto such implicit surfaces.  Several of these operations can be
viewed as energy-minimization problems that induce an instantaneous flow field
on the explicit surface. Our method uses the flow field to deform parametric
implicit surfaces by extending the classical theory of level sets. We also
derive a consolidated view for existing methods on differentiable surface
extraction and rendering, by formalizing connections to the level-set theory. We
show that these methods drift from the theory and that our approach exhibits
improvements for applications like surface smoothing, mean-curvature
flow, inverse rendering and user-defined editing on implicit geometry.
\end{abstract}

\section{Introduction}
Recent successes in generative modeling of
shapes~\cite{chen19,park19,sitzmann19meta} and inverse
rendering~\cite{niemeyer20,yariv20} are largely driven by implicit
representations of geometry parameterized as multi-layer perceptrons (MLPs) (or
neural implicits~\cite{davies2020effectiveness}). These networks can compactly represent highly-detailed
surfaces at (theoretically) infinite
resolution~\cite{martel2021acorn,sitzmann19,takikawa2021nglod,tancik2020fourier};
they are defined continuously in $\mathbb{R}^3$ and are
differentiable -- enabling their usage in gradient-based
optimization~\cite{kellnhofer2021neural,yariv20} and
learning~\cite{alldieck2021imghum,chan2021pi,mu2021sdf} methods.  Despite these
advances, there is still a large body of work in geometry processing, computer
vision and graphics which relies on explicit surface representations. Often
these mesh-based algorithms are a better choice than their implicit
counterparts. For instance, in case of inverse rendering, differentiable
renderers for triangle meshes~\cite{goel20,laine20,li18,david19,zhang20} are a)
faster, b) more accurate, and c) can handle more complex light-transport
effects, in comparison to the renderers designed for implicit
surfaces~\cite{jiang2020sdfdiff,niemeyer20,yariv20}. Similarly, geometry
processing algorithms for applications like surface smoothing and
deformation~\cite{botsch2006deformation,sorkine2007rigid,taubin1995signal} are
vastly superior in terms of compute and memory requirements than the ones
developed for neural implicit surfaces~\cite{yang21}. Most of these methods,
however, are highly specific to mesh geometry and are not easily adaptable to
MLP-defined surfaces. Our work is a theoretical attempt to bridge this gap.

We first introduce the following insight: several mesh-based algorithms define
an energy-minimization problem that is solved using gradient descent; the
gradients used by the optimizer to update the geometry can be viewed as
analogous to an instantaneous explicit flow-field ($\V$) applied on the surface. 
Informed by the literature on fluid simulation~\cite{batchelor00} and
level-sets~\cite{osher01}, deformation of a surface with such a flow field
depends on the geometry representation.

\begin{wrapfigure}[6]{r}{0.18\textwidth} \vspace{-2.1em}
 \begin{overpic}[width=0.18\textwidth, tics=15]{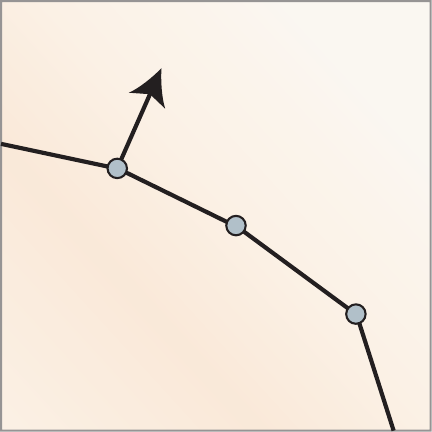} \put(4,
 6){\small{$\x \to \x'$}} \put(40, 65){\small{$\V$}} \end{overpic}
\end{wrapfigure}
The \textit{Lagrangian} representation involves tracking the surface explicitly
as a set of a points ($\x$) and connections (like triangles).  The point-set is
discrete and the connectivity is \textit{static}, which keeps the optimization
relatively simple; we can separately integrate the field at each point (update
vertices $\x \to
\x'$).  But optimization of the resolution of the surface is non-trivial and
can also get unwieldy for problems which involve surfaces with unknown
topology.

\begin{wrapfigure}[6]{l}[0pt]{0.18\textwidth}\vspace{-2.1em}
	\begin{overpic}[width=0.18\textwidth, tics=15]{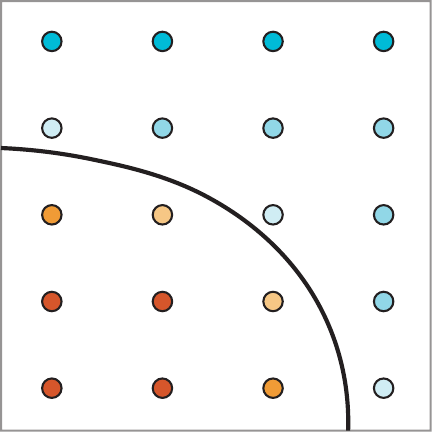} \put(2,
	9){\colorbox{white}{\small{$\phi \to \phi'$}}} \end{overpic}
\end{wrapfigure}
Alternatively, \textit{Eulerian} descriptions can be used.  Each point in space
has an object-property $\phi$ associated with it, like the distance from the
surface or its occupancy inside the enclosed volume.  The surface here is
implicitly defined with \textit{dynamic} connectivity; one can smoothly vary the
topology during optimization.  Canonically, $\phi$ is defined only on a discrete
voxel-grid and needs to be interpolated for points off the grid.  Here, making
instantaneous updates to the surface is more involved as it requires changing
$\phi$ values for a large set of points.  
A neural implicit is a continuous variant of an Eulerian representation.
Applying a flow field to such functions is non-trivial as updates are required
in the parameter ($\straighttheta$) space as opposed to directly updating $\phi$
to $\phi'$. 

We propose a parametric level-set evolution method (\S~\ref{sec:method}) which
propagates neural implicit surfaces according to an explicitly generated flow
field. Our method comprises of three repeating steps, 1) A non-differentiable
surface extraction method like Marching Cubes~\cite{lorensen87} or Sphere
Tracing~\cite{hart96} is used to obtain a Lagrangian representation corresponding
to a neural implicit, 2) A mesh-based algorithm is used to derive a flow field
on the explicit surface (\S~\ref{sec:lag}), and 3) A corresponding Eulerian flow
field is used to evolve the implicit geometry (\S~\ref{sec:eul}).  

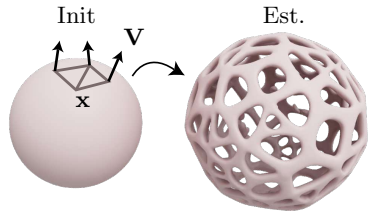
\begin{wrapfigure}[8]{r}[0pt]{0.4\textwidth}
    \vspace{-2.em}
	\begin{overpic}[width=0.4\textwidth, tics=15]{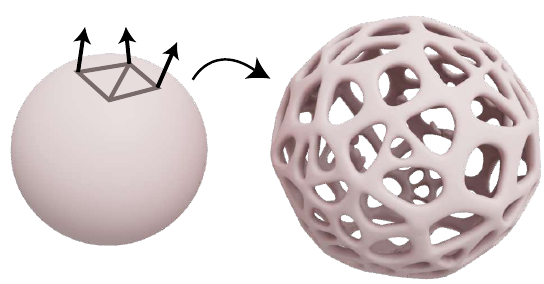} 
        \put(71,52){\small Est.}
        \put(15,52){\small Init}
        \put(33,46){\small $\V$}
        \put(20,29){\small $\x$}
    \end{overpic}
    \vspace{-2em}
    \caption{\small Inverse-rendering recovery.}
\end{wrapfigure}
Previous methods~\cite{niemeyer20,remelli20,shen2021dmtet} approach the problem of using
mesh-based energy functions on Eulerian surfaces with the idea of differentiable
extraction of Lagrangian representations. We show these methods are not in
accordance with the level-set theory~\cite{osher88} (\S~\ref{sec:theory}). The
discussion also yields a more general proof (\S~\ref{sec:meshsdf}) for
differentiable surface extraction from level set functions with arbitrary
gradient norms ($|\nabla\phi|\neq1$).  Our method is more formally connected to
the theory and we validate it with experimental observations made on three
diverse problem settings, 1) Curvature-based deformation
(\S~\ref{sec:curvature-based-deformation}), where we demonstrate more accurate
surface smoothing and mean-curvature flow than previous
methods~\cite{remelli20,yang21}, 2) Inverse rendering of geometry
(\S~\ref{sec:inverse_rendering}), where we show accurate recovery from
multi-view images for high-genus shapes without object masks as
in~\cite{niemeyer20,yariv20} (example in Fig.~1), and 3) User-defined shape editing
(\S~\ref{sec:user-editing}), where the implicit surface is deformed $60\times$
faster than previous work~\cite{yang21}.
\section{Related Work}
Coordinate-based MLPs have become a popular choice as function approximators
for signals like images, videos and
shapes~\cite{chen19,park19,sitzmann19,stanley2007cppn,tancik2020fourier}.  Our
work focuses on using MLPs for approximating implicit geometry.  Such
representations are compact, continuous and differentiable. These advantages
are well suited with gradient-descent based optimization and learning problems. Recent
developments in generative modeling of
shapes~\cite{chabra2020localshapes,mehta2021modulated,park19,sitzmann19meta},
3D consistent image synthesis~\cite{chan2021pi,xu2021volumegan}, 3D
reconstruction~\cite{azinovic2021neural,sucar2021imap,zhu2021nice} and inverse
rendering~\cite{kellnhofer2021neural,oechsle2021unisurf,yariv2021volume,yariv20},
all rely on representing geometry using MLPs.
For a more detailed discussion on recent work regarding coordinate-based representations
refer to the survey by Xie~\etal~\cite{xie2021neural}, and for inverse rendering
the survey by Tewari~\etal~\cite{tewari2021advances}.

There is also a rich literature on geometry
processing~\cite{botsch10,botsch2006deformation,botsch2007linear,sorkine2004laplacian,sorkine2007rigid} and
inverse rendering~\cite{li19cvpr,goel20,luan21,nicolet21} for explicit surface
representations. 
Yang~\etal~\cite{yang21} introduce some of the geometry processing ideas to neural
implicit surfaces, but the method could be inaccurate
(\S~\ref{sec:curvature-based-deformation}) and slow (\S~\ref{sec:user-editing}).

For inverse rendering applications, differentiable renderers for triangle meshes
are used for gradient-based optimization. Physics-based renderers
differentiate through a light-simulation
process~\cite{bangaru20,li18,david19,zhang20} and provide accurate 
gradients. Alternatively, differentiable
rasterization~\cite{laine20,liu19,ravi20} can be used for high-performance
gradient computation, but only for single-bounce rendering models. However,
optimizing triangle meshes with gradient-descent is non-trivial.  Careful design
of the optimization method~\cite{goel20,nicolet21} and error
functions~\cite{luan21} is required for robust optimization.
To circumvent some of these issues, IDR~\cite{yariv20} and DVR~\cite{niemeyer20}
use implicit surface representations like SDFs and occupancy
functions~\cite{mescheder19} parameterized with MLPs. These methods mitigate
some of the topological restrictions, but are not physics-based and are not in
accordance with the level-set theory
(\S~\ref{sec:differentiable-surface-rendering}). We propose an inverse rendering
method which uses explicit differentiable renderers for parametrically defined
implicit surfaces. The proposed method is not as sensitive to initialization as
explicit methods~\cite{goel20,nicolet21} are, does not require an object mask
like implicit methods~\cite{niemeyer20,yariv20} do, and maintains the
ability to vary topology during optimization.

Our method uses the level-set theory~\cite{osher88} as the foundation for
optimizing and deforming parametric implicit surfaces. Previous
methods~\cite{ambrosio1996level,museth2002level,whitaker1998level} for
the applications discussed in this work apply to non-parametric
level sets. Perhaps the most related works to our method are
MeshSDF~\cite{remelli20} and RtS~\cite{cole2021differentiable}. Compared
to MeshSDF, our approach is more formally connected to the theory of
level-sets (\S~\ref{sec:method}), applies to all parametric functions
 (\S~\ref{sec:meshsdf}), and works for a
more diverse set of optimization problems like shape editing
(\S~\ref{sec:curvature-based-deformation}, \ref{sec:user-editing}) and
inverse rendering (\S~\ref{sec:inverse_rendering}) -- deviating from experimental
observations made by Remelli~\etal~\cite{remelli20} on
learning-based settings. Compared to RtS~\cite{cole2021differentiable},
we show geometry processing applications along with theoretical
parallels (\S~\ref{sec:theory}) between parametric level-set methods like
MeshSDF~\cite{remelli20}, DVR~\cite{niemeyer20}, IDR~\cite{yariv20} and
the classical theory~\cite{osher01}.  Our inverse rendering method is
shown to work (\S~\ref{sec:inverse_rendering}) for a set of high-genus
shapes with a genus-0 initialization, in contrast to object-pose
optimization or small genus changes shown
in~\cite{cole2021differentiable}. Recent work by Munkberg
\etal~\cite{munkberg2021} and Hasselgren \etal~\cite{hasselgren2022shape}
also show promise in using explicit differentiable renderers with
implicit geometry for inverse problems.

\section{Background}
Consider a closed surface of arbitrary topology $\partial\Omega$ evolving with
respect to time $t$. We define a Lagrangian representation of this surface with a finite set of $k$
points in $\mathbb{R}^3$ as $\partial\Omega_L = \{\x_i\ |\ \x_i \sim \partial\Omega;\ \forall i
\in\{1, 2, 3, \dots, k\}\}$. This point-set can be viewed as a triangle mesh if an additional set of
connections between the points is provided, and as a point cloud otherwise. Implicitly this surface
can also be represented with a family of level-sets $\phi: \mathbb{R}^{3} \to \mathbb{R}$, the zero
iso-contour of which represents the surface $\partial\Omega_E = \{\x\ |\ \phi(\x) = 0\}$.  $\phi$
can be arbitrarily chosen, but a canonical choice is a signed-distance function (SDF) which
satisfies:
\begin{align}
\phi(\x) = (\pm)\min_{\x_C \in \partial\Omega}\{||\x -
\x_C||_2\},\label{eq:sdf}
\end{align}
\\
where $\x_C$ is the closest point on the surface to $\x$ and the sign of
$\phi(\x)$ denotes whether $\x$ is enclosed ($-$) within the shape or not
($+$).  
\paragraph{Parameterizing $\phi$} Analytically defining $\phi$ for simple and
regular shapes is relatively straightforward~\cite{quilez}, but is infeasible
for most objects.  Recent work on 3D reconstruction~\cite{niemeyer20,yariv20}
and generative shape modeling~\cite{chen19,park19} suggests parameterizing
$\phi$ using a multi-layer perceptron (MLP) with $\straighttheta$ as its
parameters.  The networks are optimized by minimizing an energy function
comprised of a distance term~\cite{park19} and a gradient term~\cite{gropp20}
enforcing $|\nabla\phi|$ to be $1$.  We use SIREN~\cite{sitzmann19} as the
parametric function of choice, although our method is agnostic to the
network parameterization. The network acts approximately as an SDF at the rest state ($t
= 0$), but may not retain the SDF property (Eq.~\ref{eq:sdf}) as the surface
evolves. We denote this surface as $\partial\Omega_E($\straighttheta$) =\{ \x\
|\ \pphi(\x; \boldsymbol{\straighttheta}) = 0\}$, where $\pphi$ is parameterized
with $\straighttheta$ as the weights and biases of the network. For clarity, we
use $\pphi$ for parametric level-sets and $\phi$ for non-parametric.

\section{Method}
\label{sec:method}
\begin{figure}[t!]
    \centering
    \begin{overpic}[width=0.98\linewidth, tics=5]{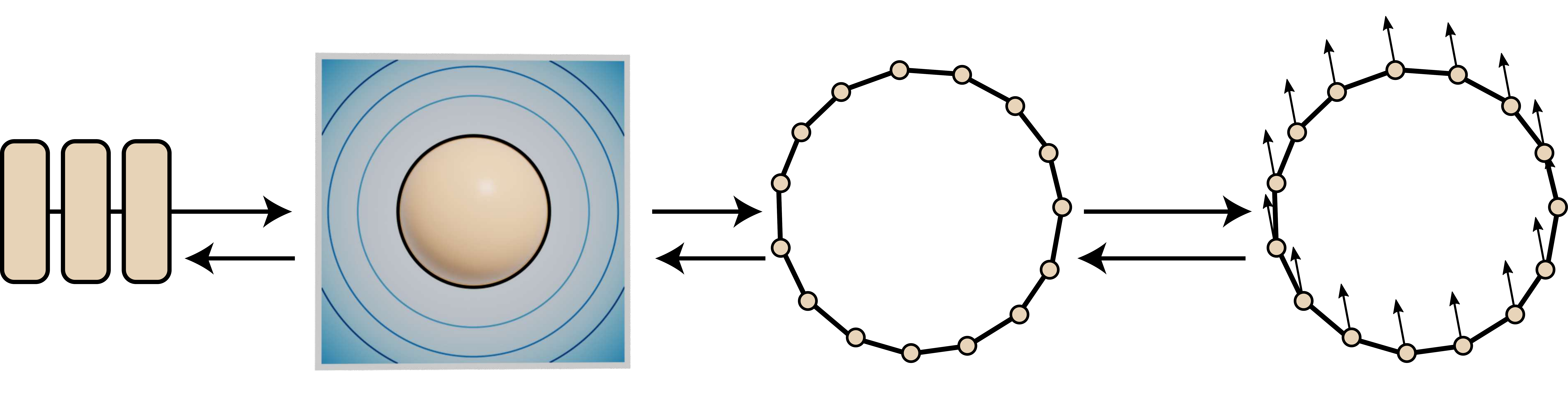}
        \put(5, 21) {\makebox[0pt]{\small Parametric}}
        \put(5, 18) {\makebox[0pt]{\small Level-Sets}}
        \put(30, 23) {\makebox[0pt]{\small Eulerian $\partial\Omega_E$}}
        \put(59, 23) {\makebox[0pt]{\small Lagrangian $\partial\Omega_L$}}
        \put(45, 14) {\makebox[0pt]{\small MC}}
        \put(45, 4) {\makebox[0pt]{$\frac{\partial\phi}{\partial t}$}}
        \put(15.5, 14) {\makebox[0pt]{\small $\pphi(\x;\straighttheta)$}}
        \put(74, 14) {\makebox[0pt]{\small $\min_{\x_i} \mathcal{E}$}}
        \put(74, 4) {\makebox[0pt]{$\frac{d\x}{dt}$}}
        \put(66, 20) {\makebox[0pt]{\small $\x_i$}}
        \put(100, 20) {\makebox[0pt]{\small $\V(\x_i)$}}
        \put(15, 4) {\makebox[0pt]{\small Update}}
        \put(15, 1.5) {\makebox[0pt]{\small $\straighttheta$}}
    \end{overpic}
    \caption{\label{fig:main}
        \textbf{Method Overview.} We present a level-set method to evolve neural
        representations of implicit surfaces. Using Marching Cubes
        (MC)~\cite{lorensen87}, a Lagrangian surface $\partial\Omega_L$ is
        extracted from an Eulerian representation $\partial\Omega_E$ encoded in
        the network parameters $\straighttheta$.  An energy function $\mathcal{
        E}$ is defined
        on $\partial\Omega_L$ which is minimized using gradient-descent. The
        gradients of the optimizer together act as a flow-field $\V$ on the surface
        points $\x$, which is used to evolve the non-parametric $\phi$ using
        the level-set equation. The values of $\phi$ on the surface
        act as references to update the parameters $\straighttheta$ of the
        network.
    }
\end{figure}
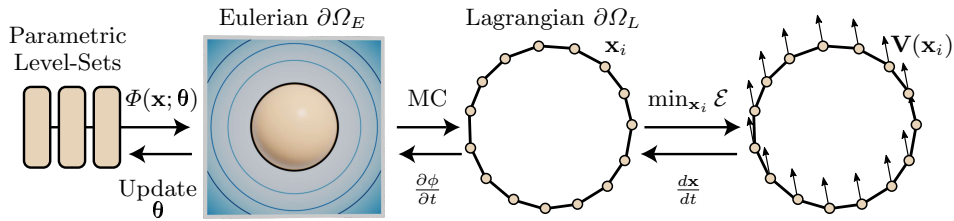

We begin the discussion by first characterizing the deformation of surfaces into
Lagrangian (\S~\ref{sec:lag}) and Eulerian (\S~\ref{sec:eul}) settings. We show
that gradient descent on energy functions defined for triangle meshes can be
viewed as surface deformation under the dynamics of a flow field $\V$, which is
\textit{discretely} defined only on the surface points $\partial\Omega_L$.
We can use this flow field to deform a \textit{continuous} surface
representation using the level-set equation. We extend the
level-set equation to the case of parametric level-sets $\pphi$
which enables us to use loss functions defined on triangle meshes to deform
MLP-defined level-sets.

\subsection{Lagrangian Deformation}
\label{sec:lag}
As mentioned earlier, in the Lagrangian setting, the surface is defined with a
finite set of points $\partial\Omega_L$.  A variety of methods in
geometry processing and computer vision define an energy function $\mathcal{E}$ that is
minimized to make instantaneous updates to $\partial\Omega_L$. Some recent
examples include optimizing point-clouds for
view-synthesis~\cite{aliev20,ruckert21} and triangle-meshes for
inverse-rendering of geometry~\cite{goel20,nicolet21}. The surface is updated
using spatial gradients $\frac{\partial \mathcal{E}}{\partial \x}$, which is well
studied in numerical analysis~\cite{suli03} and optimization
methods~\cite{chong04}. Through the lens of physics, these gradients induce an
instantaneous flow field $\V(\x)$, which can be used to evolve the surface by
integrating the following ordinary differential equation (ODE):
\begin{align}
\frac{d\x}{d t} = -\frac{\partial \mathcal{E}}{\partial \x} \rightarrow\V(\x).\qquad\lhd\ \text{Lagrangian Deformation}  \label{eq:lag}
\end{align}
Numerically, this can be done using forward-Euler steps $\x^{t+1} =
\x^t + \Delta t  \V^t(\x)$. This is easy to accomplish if the connectivity
of the points remains static. More sophisticated integration schemes can also be
used~\cite{suli03}.  Here, in case of optimization problems
solved using gradient descent, $\Delta t$ is equivalent to the learning rate. Several works
in shape deformation~\cite{gupta20} and inverse rendering~\cite{nicolet21} can
be subsumed by this ODE with different definitions for flow $\V$ and time-step
$\Delta t$. We next show that these readily available energy functions and
optimization algorithms defined for explicit surfaces can also be used to
optimize MLP-defined level-sets.

\subsection{Eulerian Deformation} 
\label{sec:eul}
\label{sec:eul}
To avoid the topological complications associated with Lagrangian deformations,
we can instead define a corresponding Eulerian deformation field. By definition,
we know for points $\x \in \partial\Omega$, $\phi(\x) = 0$. Using implicit
differentiation:
\begin{align}
&\frac{d \phi(\x)}{dt} = 
\frac{\partial \phi}{\partial t} + \frac{\partial\phi}{\partial
\x}\frac{\partial \x}{\partial t} 
=\frac{\partial \phi}{\partial t} + \nabla \phi\cdot \V = 0 \qquad
\lhd\ \text{From~\ref{eq:lag}}\nonumber\\
&\iff \frac{\partial \phi}{\partial t} = -\nabla \phi\cdot
\V.\qquad\lhd\ \text{Eulerian Deformation} \label{eq:eul}
\end{align}
This partial differential equation (PDE) is sometimes referred to as the
level-set equation~\cite{osher88}, the material derivative~\cite{batchelor00} or
the G-equation~\cite{markstein14}.  We extend this PDE to obtain an evolution
method for parametric level-sets $\pphi$. First, for each time step $t$, we
extract a Lagrangian surface representation $\partial\Omega_L^t$ from $\pphi$
using MC~\cite{lorensen87}. Depending on the task at hand, an energy
function $\mathcal{E}$ (\eg, photometric error for inverse rendering) is defined on
$\partial\Omega_L^t$. Assuming $\mathcal{E}$ is differentiable, we compute
$\frac{\partial \mathcal{E}}{\partial\x} = -\V^t(\x)$ for each vertex $\x_i \in
\partial\Omega_L$. With the flow field $\V^t$, we update the level-set function
as we would in the non-parametric case using forward-Euler steps:
\begin{align}
\phi^{t+1} = \pphi^{t} - \Delta t \nabla\pphi^{t}\cdot\V^t. \qquad\lhd\
\text{From (\ref{eq:eul})}\label{eq:eul_d}
\end{align}
The time step $\Delta t$ here is a parameter which is dependent on the dynamics of the
flow-field. If $\V$ is highly non-linear, taking smaller steps (\ie, $\Delta t$
is small) is required, while for a simple field, larger values of $\Delta t$
should suffice. For each time step, we take the values of non-parametric
$\phi^{t+1}$ as the reference and update the parameters of 
$\pphi$ accordingly. This is achieved by minimizing the following objective: \begin{align}
    \min_{\straighttheta} J(\straighttheta) = \frac{1}{|\partial\Omega_L|}\sum_{\x \in \partial\Omega_L} ||\phi^{t+1}(\x) - \pphi(\x;
    \straighttheta)||^2\
    ,\label{eq:J}
\end{align}
using gradient descent. Since the surface updates are small for each time step,
the number of descent steps required is in the order of $10^2$. This makes the
method convenient for obtaining neural representations of deformed variants of
the initial geometry. After each optimization routine, we again extract the
Lagrangian surface using MC~\cite{lorensen87}, which is subsequently
fed into a mesh-based energy-minimization problem. An overview of our method is
shown in Figure~\ref{fig:main}. For each of the applications we show in
\S~\ref{sec:applications}, we define $\V^t$ using a corresponding energy function,
and minimize $J$ for each time step.

\section{Theoretical Comparisons}
\label{sec:theory}
The level-set method discussed in \S~\ref{sec:eul} subsumes two related works,
1) Differentiable iso-surface extraction (MeshSDF)~\cite{remelli20}, and
2) Differentiable rendering of implicit surfaces
(DVR/IDR)~\cite{niemeyer20,yariv20}. We first show
 that MeshSDF minimizes the level-set objective
$J$ defined in (\ref{eq:J}) with a single gradient-descent step. But the surface
does not propagate in agreement with the level-set equation, as outlined in
\S~\ref{sec:meshsdf}.  We then show that these two seemingly disparate works
(MeshSDF and DVR) are closely related in
\S~\ref{sec:differentiable-surface-rendering}. We end the discussion with an
explanation for how DVR deviates from the level-set equation.
\subsection{Differentiable Iso-Surface Extraction}
\label{sec:meshsdf}
\begin{result}
    \label{result:single_step}
    Differentiable Iso-Surface Extraction~\cite{remelli20} takes a single
    gradient-descent step to minimize the level-set objective function
    $J$ (Equation~\ref{eq:J}).
\end{result}

    \begin{wrapfigure}[15]{r}[0pt]{0.5\textwidth} 
    \vspace{-2.1em}
    \begin{overpic}[width=0.5\textwidth, tics=15]{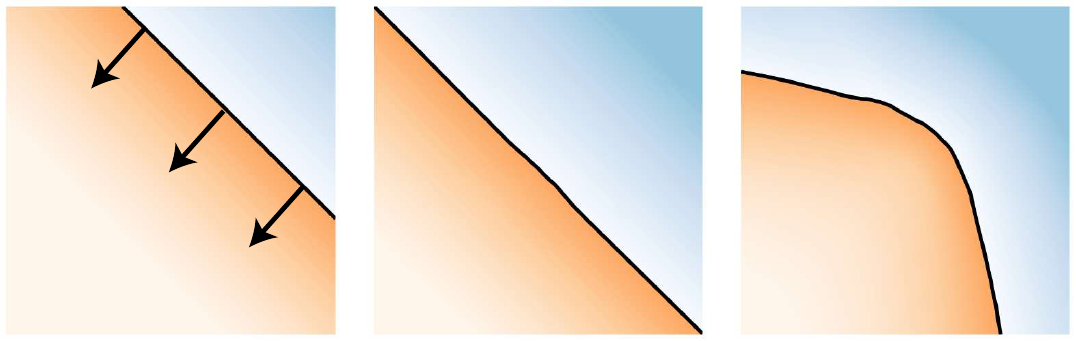}
        \put(11, 33){\small{Init}}
        \put(44, 33){\small{Ours}}
        \put(71, 33){\small{MeshSDF$^\straighttheta$}}
    \end{overpic}
    \caption{\label{fig:single_step}\textbf{MeshSDF$^\straighttheta$ does not follow the
        level-set equation}. (\textit{Left}) A planar surface defined implicitly
        with an MLP is influenced by a flow field in the direction of its
        normal.  The motion attained using differentiable iso-surface extraction
        (\textit{Right}) is inconsistent with the field. Our method
        (\textit{Center}) propagates the front as expected.}
\end{wrapfigure}

    \noindent \textit{Proof.} MeshSDF~\cite{remelli20} defines a loss-function
    $\mathcal{L}$ on a triangle mesh extracted using Marching
    Cubes~\cite{lorensen87} from an SDF parameterized with an MLP. They use an
    MLP $\pphi(\x; \straighttheta,\z)$ conditioned on a latent-code $\z$
    characterizing the shape. Using $\mathcal{L}$ they update the latent-code
    $\z$, which is different from our goal of updating $\straighttheta$ for an
    unconditional $\pphi$. To clarify this distinction, we use
    MeshSDF$^{\straighttheta}$ to denote our variant which updates
    $\straighttheta$. To update the parameters of the MLP, we compute the following
    gradient using the chain-rule: 
    \begin{align}
    \frac{\partial\mathcal{L}}{\partial\straighttheta} =
    \sum_{\x\in\partial\Omega_L}
    \frac{\partial\mathcal{L}}{\partial\x}\frac{\partial\x}{\partial\pphi}\frac{\partial\pphi}{\partial\straighttheta},
    \label{eq:meshsdf_1}
    \end{align}
    where $\x$ are the vertices on the mesh and $\pphi$ is an SDF. The first and
    the third gradient terms on the right are computed using automatic
    differentiation. The second term $\frac{\partial\x}{\partial\pphi}$ can be
    approximated as the inverted surface normal $-\n(\x) = -\nabla_{\x}\pphi$~\cite{remelli20}, when $\pphi$ is an SDF.
    In the spirit of Lagrangian deformation (Equation~\ref{eq:lag}),
    $-\frac{\partial\mathcal{L}}{\partial\x}$ acts as an instantaneous flow field
    $\V$ on the vertices $\x$. The parameters of the MLP are then updated
    as:
    \begin{align}
        \straighttheta \leftarrow \straighttheta - \lambda \frac{\partial
        \mathcal{L}}{\partial\straighttheta} = \straighttheta - \lambda\sum_{\x\in\partial\Omega_L}
        \V\cdot\nabla\pphi\frac{\partial\pphi}{\partial\straighttheta},
        \label{eq:meshsdf_2}
    \end{align}
    where $\lambda$ is the learning rate. Alternatively, we can also update
    $\straighttheta$ using the objective function defined in Equation~\ref{eq:J} using
    gradient descent:
    \begin{align}
        \straighttheta \leftarrow \straighttheta - \lambda \frac{\partial
        J}{\partial\straighttheta} &= \straighttheta - \lambda\sum_{\x \in
    \partial\Omega_L} 2(\phi^{t+1}(\x) -
        \pphi(\x;\straighttheta))\left(-\frac{\partial\pphi}{\partial\straighttheta}\right)\nonumber\\
                           &= \straighttheta -
        \epsilon\sum_{\x\in\partial\Omega_L}\V\cdot\nabla\pphi\frac{\partial\pphi}{\partial\straighttheta},
        \quad\lhd\ \text{From (\ref{eq:eul_d})} \label{eq:single_step}
    \end{align}
    where $\epsilon$ is a constant. The last equivalency is valid when
    $\phi^{t+1}(\x) - \pphi(\x;\straighttheta) = -\Delta t\V\cdot\nabla\pphi$
    (Eq.~\ref{eq:eul_d}),
    which is true only for the first gradient descent step. Subsequently, the
    value of $\pphi(\x; \straighttheta)$ changes as the parameters get updated.
    Comparing (\ref{eq:meshsdf_2}) and (\ref{eq:single_step}), we conclude that
    the optimization in MeshSDF$^{\straighttheta}$ has the effect of taking a single
    gradient-descent step to minimize $J$. Note that while the proof by
    Remelli~\etal~\cite{remelli20} assumes $\pphi$ is an SDF, the second update
    equation (\ref{eq:single_step}) does not.  It is valid for all level-set
    functions, with no restrictions on the values of the gradient-norm
    ($|\nabla\pphi|$) and is also valid for occupancy functions~\cite{mescheder19}. $\square$

However, by taking a single step to update $\straighttheta$, the surface does
not propagate in agreement with the level-set equation (\ref{eq:eul}). This is
problematic for applications which require the surface to move as intended by the
flow-field. An example application is shown in
\S~\ref{sec:curvature-based-deformation}. We also illustrate this with a toy
example in Figure~\ref{fig:single_step} where a planar surface propagates in the
direction of its normal. A more formal discussion is in
Result~\ref{result:normal}.

\begin{result}
    \label{result:normal}
    Dfferentiable iso-surface extraction~\cite{remelli20} does not propagate the
    surface-front as dictated by the flow field.
\end{result}
\begin{proof}
    We show this with an example flow field. Consider the surface-front
    propagating with a constant speed $\beta$ in the direction of the normal.
    The corresponding Eulerian deformation is constant across the surface:
    \begin{align}
        \frac{\partial\phi}{\partial t} = -|\nabla\phi|\beta = -\beta.
        \qquad\lhd\ \text{Assuming $\phi$ is an SDF}\label{eq:const_speed}
    \end{align}
    With the same flow field, we can estimate the instantaneous change in $\pphi$
    (parametric) for MeshSDF$^\straighttheta$ as:
    \begin{align}
        \frac{\partial\pphi}{\partial t} =
        \frac{\partial\pphi}{\partial\straighttheta}\frac{\partial\straighttheta}{\partial
        t} =
        \frac{\partial\pphi}{\partial\straighttheta}\sum_{\x\in\partial\Omega_L}
        -\beta\frac{\partial\pphi}{\partial\straighttheta} =
        \frac{\partial\pphi}{\partial\straighttheta}B. \quad\lhd\ \text{From
            (\ref{eq:meshsdf_2}) and (\ref{eq:const_speed})}
    \end{align}
    The term $B$ on the right is constant for every point $\x$ on the surface.
    The gradient $\frac{\partial\pphi}{\partial\straighttheta}$ is dependent on
    the position where it is evaluated and hence front-propagation
    $\frac{\partial\pphi}{\partial t}$ is not constant. On the contrary,
    we minimize the objective function defined in (\ref{eq:J}) which ensures that
    the surface propagation is constant as in (\ref{eq:const_speed}).\ $\square$
\end{proof}
For the implications of Result~\ref{result:single_step} and \ref{result:normal}, and
experimental comparisons, we defer the discussion to
\S~\ref{sec:curvature-based-deformation}.

\subsection{Differentiable Surface Rendering}
\label{sec:differentiable-surface-rendering}
An alternate way of extracting a Lagrangian surface  $\partial\Omega_L$ from
$\pphi$ is by computing ray-surface intersections using ray-marching or
sphere-tracing~\cite{hart96}. If ray-marching is
differentiable~\cite{jiang20,liu20}, one can backpropagate gradients from error
functions defined on $\partial\Omega_L$ to the parameters $\straighttheta$
defining the implicit surface.  Recent developments in inverse
rendering~\cite{mescheder19,yariv20} rely on this idea. The explicit surface
extracted using ray-marching differs from the one obtained using marching-cubes
in two ways, 1) Ray-marching does not extract the
connectivity (\eg, triangle faces) among the intersection points. This restricts the
usage of loss functions which rely on attributes like the edge-length or
differential operators like the Laplacian. 2) The intersection points depend on
the camera attributes. The resolution of the image plane affects the density of
points and the viewing direction determines the visibility. As a result,
$\partial\Omega_L$ obtained using ray-marching could be sparser than the one
obtained marching cubes.
Furthermore, as we show in
Result~\ref{result:dvr_meshsdf}, by using differentiable ray-marching the
parameters get updated exactly as when differentiable iso-surface extraction
is used---although with
a less favorable Lagrangian representation (sparser and no connectivity). This is in addition to the
computational disadvantage associated with ray-marching. We also formally show
in Result~\ref{result:dvr_lse} that surface evolution with differentiable
ray-marching is in disagreement with level-set theory for tangential flows.

\begin{result}
    \label{result:dvr_meshsdf}
    Surface evolution using differentiable ray-marching of parametric implicit
    surfaces~\cite{mescheder19,yariv20} is the same as using differentiable
    iso-surface extraction~\cite{remelli20} when the viewing direction
    \textnormal{$\textbf{v}_u$} is
    parallel to the normal \textnormal{$\n$} at the intersection point \textnormal{$\x_u$}. The parameters $\straighttheta$
    for the level-set function $\pphi$ are updated as:
\end{result}
\begin{align}
    \straighttheta \leftarrow \straighttheta - \lambda \sum_{\x_u} \V\cdot
    \nabla\pphi\frac{\partial\pphi}{\partial\straighttheta},
    \label{eq:dvr_meshsdf}
\end{align}
where $\x_u$ are the visible points and $\V$ is the flow field. Comparing
(\ref{eq:dvr_meshsdf}) and (\ref{eq:meshsdf_2}), the gradient-descent step is
the same. As in the case of MeshSDF$^\straighttheta$, here the surface is
evolved with a single step in the parameter space. We provide a more detailed
proof for (\ref{eq:dvr_meshsdf}) in the Appendix.

\begin{result}
    \label{result:dvr_lse}
    Differentiable ray-marching of parametric implicit
    surfaces~\cite{mescheder19,yariv20} disagrees with the level-set equation for
    tangential components \textnormal{$\V^\perp$} of the flow field
    \textnormal{$\V$}. The change in parameters
    $\Delta\straighttheta$ is:
\end{result}
\begin{align}
   \Delta\straighttheta = \lambda\sum_{\x_u}\pm|  \V^\perp
   | \tan (\arccos (\nabla\pphi \cdot
   \textbf{v}_u))\frac{\partial\pphi}{\partial\straighttheta}
   \neq 0,\label{eq:dvr_tangent}
\end{align}
\begin{wrapfigure}[16]{r}{0.7\textwidth} 
    \vspace{-1.em}
    \begin{overpic}[width=0.7\textwidth, tics=15]{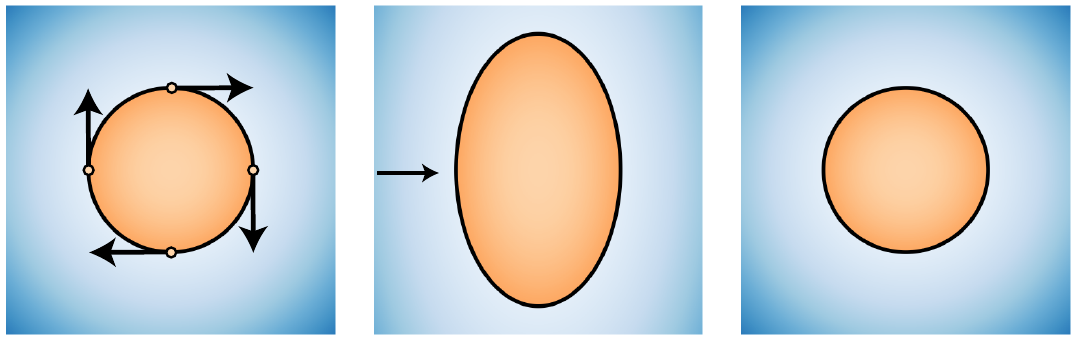}
        \put(13, 25){\small $\x_u$}
        \put(23, 25){\small $\V^\perp$}
        \put(35, 17){\small $\textbf{v}_u$}
        \put(13, 33){\small Init}
        \put(35, 33){\small DVR/IDR~\cite{mescheder19,yariv20}}
        \put(80, 33){\small Ours}
    \end{overpic}
    \caption{\label{fig:dvr}\textbf{Tangential flow fields may deform
    surfaces when DVR/IDR is used for surface extraction.} (\textit{Left}) A
    parametric Eulerian circle undergoes tangential deformation $\V^\perp$ at surface
    points $\x_u$.  (\textit{Middle}) Using differentiable surface rendering,
    the surface deforms incorrectly. (\textit{Right}) Our method agrees with the
level-set equation and the resultant deformation is the identity.}
\end{wrapfigure}
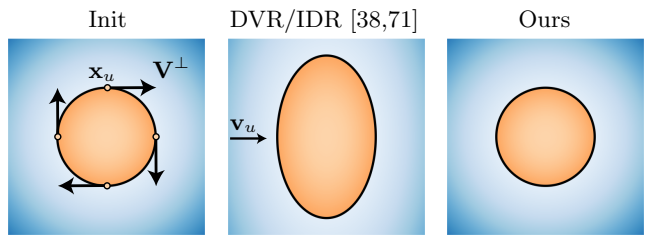

\hspace{-1.5em}which could be non-zero. A detailed proof is in the Appendix.
Referring to Eq.~\ref{eq:eul}, for tangential flows the
surface should not undergo any deformation, i.e.  $\frac{\partial\phi}{\partial
t} = 0$. However, since $\straighttheta$ gets updated as in (\ref{eq:dvr_tangent}), the
surface \textit{does} deform. 
We instead minimize
the objective function $J$ (Eq.~\ref{eq:J}) which is $0$ for a tangential
field. We show an example of tangential deformation in Figure~\ref{fig:dvr}.
Experimental comparisons in \S~\ref{sec:inverse_rendering} validate that
deforming $\pphi$ accurately is critical for applications like inverse
rendering.

\section{Applications}
\label{sec:applications}
We focus on validating the proposed theory with computer
graphics models in three different settings, 1) Curvature-based
deformations (\S~\ref{sec:curvature-based-deformation}), which can be used to smooth/sharpen features and apply curvature
defined flows on implicit surfaces, 2) Inverse rendering of geometry
(\S~\ref{sec:inverse_rendering}), where a
differentiable renderer for explicit geometry can be used to evolve implicit
surfaces, and 3) User-defined deformations (\S~\ref{sec:user-editing}), where a user can specify alterations
for a given object.

\subsection{Curvature-based Deformation}
\label{sec:curvature-based-deformation}
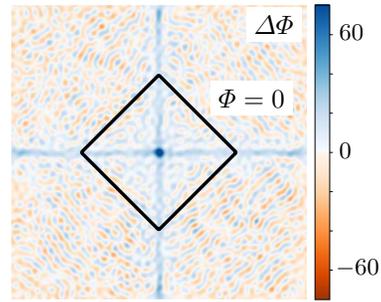
\begin{wrapfigure}[19]{r}[-4pt]{0.35\textwidth}
    \vspace{-1.5em}
    \begin{overpic}[width=0.35\textwidth, tics=15]{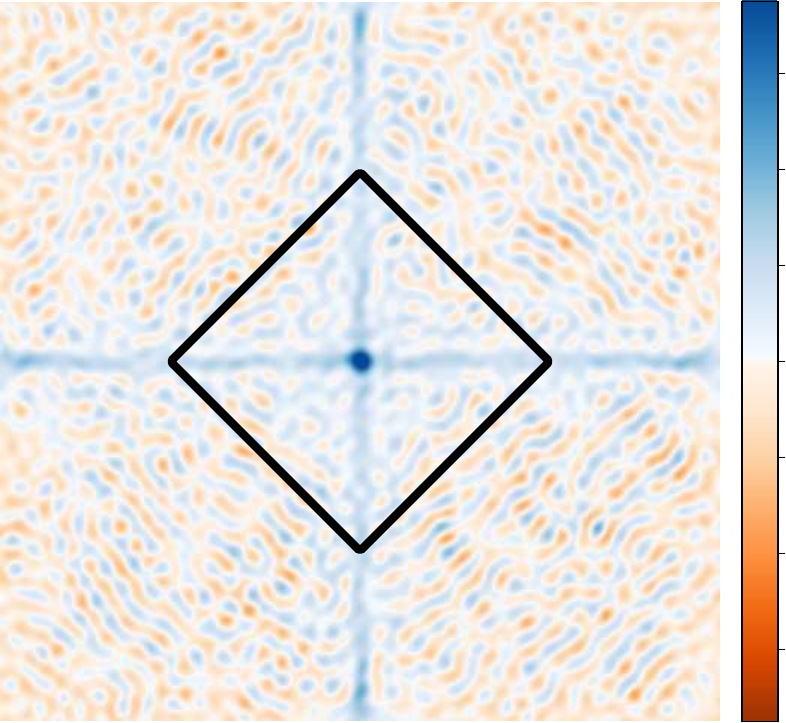}
        \put(62, 60){\colorbox{white}{$\pphi = 0$}}
        \put(73, 83){\colorbox{white}{$\Delta\pphi$}}
        \put(102, 44){\small $0$}
        \put(102, 81){\small $60$}
        \put(101, 8){\small $-60$}
    \end{overpic}
    \caption{\label{fig:laplacian}\textbf{The laplacian $\Delta\pphi$ of an
        MLP-defined level-set function is noisy}. We show the
        mean-curvature values for a parametric level-set function of a square.
        Large values are observed for a zero-curvature surface.}
\end{wrapfigure}

\begin{figure}[t]
    \centering
    \begin{overpic}[width=\linewidth, tics=5]{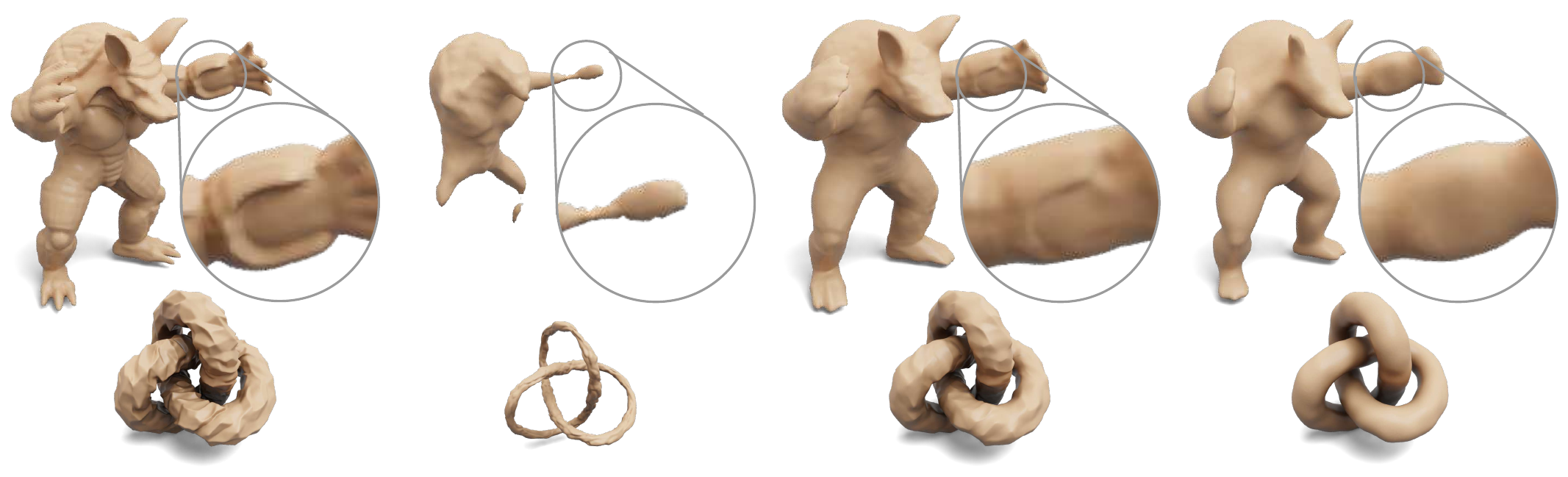}
        \put(13, -2) {\makebox[0pt]{Init}}
        \put(36, -2) {\makebox[0pt]{MeshSDF$^\straighttheta$~\cite{remelli20}}}
        \put(62, -2) {\makebox[0pt]{NFGP~\cite{yang21}}}
        \put(87, -2) {\makebox[0pt]{Ours}}
    \end{overpic}
	\vspace{-0.8em}
    \caption{\label{fig:smooth}\textbf{Surface smoothing on parametric
        level-sets.} We apply surface smoothing on an MLP-defined implicit
        surface by minimizing Dirichlet energy on the corresponding explicit
        surface. We use a discrete Laplacian to define a flow-field on the
    surface; NFGP~\cite{yang21} uses its continuous counterpart and preserves
too many high-frequency details. MeshSDF$^\straighttheta$~\cite{remelli20} fails to smoothen the surface.}
\end{figure}

To apply surface smoothing on parametric implicit surfaces, we first define a corresponding explicit
 force field. For the extracted Lagrangian surface $\partial\Omega_L$
at each time step, we minimize the Dirichlet-energy functional on the surface.
In the continuous setting, this is defined as:
\begin{align}
    \mathcal{E}(\partial\Omega) = \int_{\partial\Omega} ||\nabla \textbf{x}||^2\ d\textbf{x}.
\end{align}
Minimizing $\mathcal{E}$ can be shown~\cite{botsch10} to induce the following
explicit flow-field
on the surface:
\begin{align}
\frac{\partial\textbf{x}}{\partial t} =
\V(\textbf{x}) = \lambda\Delta\textbf{x} =  
-2\lambda\kappa(\x)\textbf{n}(\x),
\label{eq:mean_flow}
\end{align}
where $\lambda$ is a scalar diffusion coefficient, $\Delta$ is the
Laplace-Beltrami operator and $\textbf{n}$ is the normal.
$\kappa$ is the mean-curvature, which for an implicit surface is defined as the divergence of the
normalized gradient of $\phi$ (\ie, $ \nabla \cdot
\frac{\nabla\phi}{|\nabla\phi|}$)~\cite{osher01}. It is equivalent to computing
the laplacian $\Delta\pphi$ of the MLP using automatic differentiation. In
Figure~\ref{fig:laplacian} we show that such an estimation of the laplacian is
noisy---significant magnitudes are observed even for surfaces with zero
curvature.\footnote{This might be due to the unconstrained Lipschitz constants
of MLPs~\cite{liu22}.} 
Instead of using (\ref{eq:mean_flow}) as it is, which requires estimating $\Delta\pphi$, we approximate
$\V = \lambda\Delta\x \approx \lambda\textbf{L}\x$; where $\textbf{L}$ is the
discrete Laplacian we can compute using the Lagrangian surface. Note that this
is
feasible only because of the hybrid (Eulerian+Lagrangian) nature of our method. We use the
flow-field to update $\pphi$ using the method outlined in \S~\ref{sec:eul}.  Figure~\ref{fig:smooth}
shows qualitative comparisons for smoothing applied on two surfaces.  We
show a comparison with a method which applies deformation using the continuous
Laplacian ($\Delta\pphi$) (NFGP)~\cite{yang21}. When $\mathcal{E}$ is minimized using
MeshSDF$^\straighttheta$~\cite{remelli20}, the deformation is not curvature based and
high-frequency features are retained during the evolution.

\begin{figure}[b!]
    \centering
    \begin{overpic}[width=0.7\linewidth, tics=5]{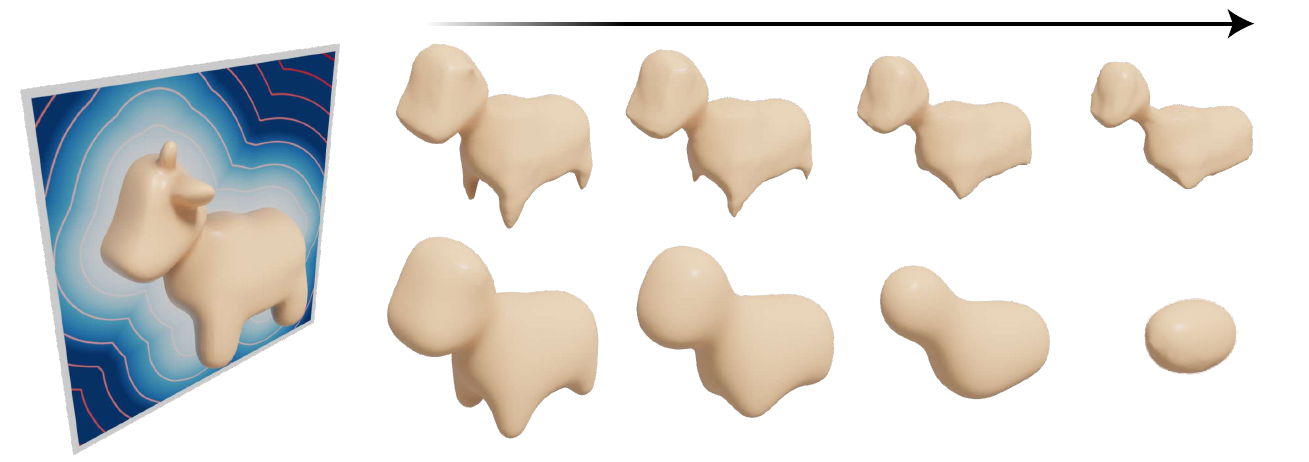}
        \put(50, 35){Mean-Curvature Flow}
        \put(100, 16){\rotatebox{90}{MeshSDF$^\straighttheta$}}
        \put(100, 5){\rotatebox{90}{Ours}}
    \end{overpic}
    \caption{\label{fig:mcf}
        We use an explicit mean-curvature flow-field to deform a parametrically
        defined implicit surface. When the same flow-field is used with MeshSDF,
        the deformation is not curvature-based. A genus-0 surface morphs into a
        sphere using our method while MeshSDF retains high-curvature regions.
    }
\end{figure}

Equation (\ref{eq:mean_flow}) is referred to as mean-curvature
flow~\cite{desbrun99}.  Since our method deforms the surface in accordance with
the flow-field, we can use (\ref{eq:mean_flow}) to apply mean-curvature flow on
a parametrically defined $\phi$. Yang~\etal~\cite{yang21} minimize an objective
function which is handcrafted for a specific level-of-smoothness. Applying
curvature-based flow is infeasible with their method since it would require
a new optimization objective for each level-of-smoothness.
As MeshSDF$^\straighttheta$~\cite{remelli20} does not evolve the surface
according to the level-set equation, the flow obtained with it is incorrect. We
show an example flow on a genus-0 surface in Figure~\ref{fig:mcf}.

\subsection{Inverse Rendering of Geometry}
\label{sec:inverse_rendering}
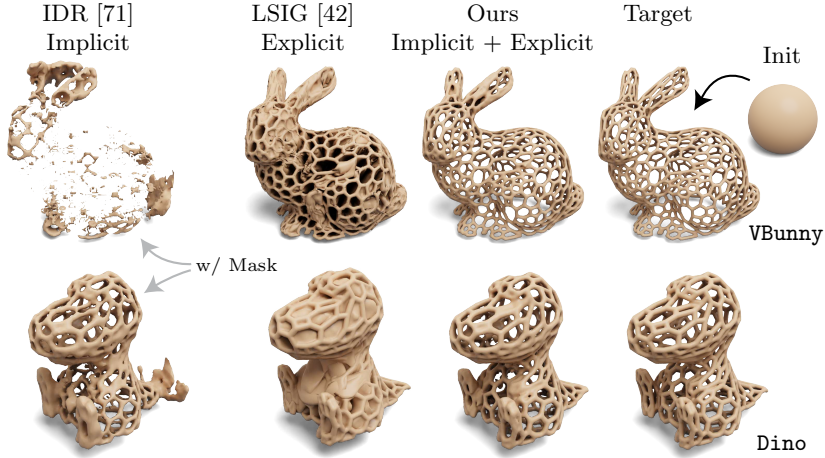
\begin{figure}[t]
    \vspace{1em}
    \centering
    \begin{overpic}[width=0.9\linewidth, tics=5]{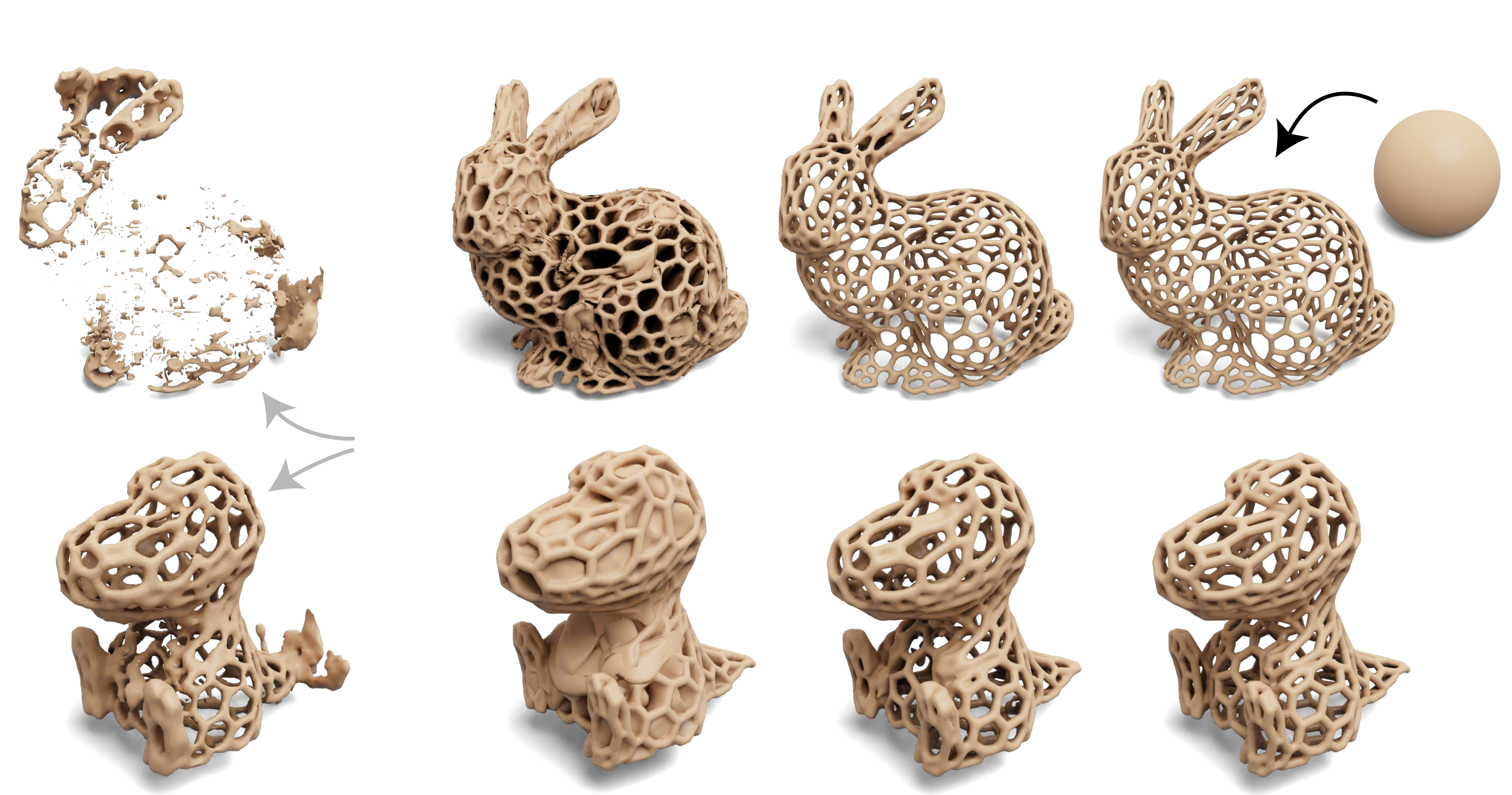}
        \put(11, 53) {\makebox[0pt]{\small IDR~\cite{yariv20}}}
        \put(37, 53) {\makebox[0pt]{\small LSIG~\cite{nicolet21}}}
        \put(60, 53) {\makebox[0pt]{\small Ours}}
        \put(80, 53) {\makebox[0pt]{\small Target}}
        \put(95, 48) {\makebox[0pt]{\small Init}}
        \put(95.5, 26.5) {\makebox[0pt]{\small \texttt{VBunny}}}
        \put(95, 1) {\makebox[0pt]{\small \texttt{Dino}}}
        \put(11, 49.5) {\makebox[0pt]{\small Implicit}}
        \put(37, 49.5) {\makebox[0pt]{\small Explicit}}
        \put(60, 49.5) {\makebox[0pt]{\small Implicit + Explicit}}

        \put(24, 22.75) {\scriptsize w/ Mask}
    \end{overpic}
    \caption{\label{fig:inverse_main}
        \textbf{Inverse rendering of high-genus shapes.} A
        spherical surface is used for initialization. IDR~\cite{yariv20}, which
        uses differentiable rendering of implicit surfaces does not recover
        finer details. LSIG~\cite{nicolet21} uses a triangle-mesh 
        and restricts the topology post initialization. Using an explicit
        differentiable renderer to optimize implicit geometry, our method can change
        topology during optimization and recover fine-details. Note that IDR
        requires an object mask and a neural renderer.
    }
\end{figure}
 
\begin{table}[t]\centering
    \includegraphics[width=\linewidth]{assets/fig20}
    \resizebox{\columnwidth}{!}{
    \begin{tabular}{@{}r@{\hskip 1em}cccccccc}
    \toprule
    \minibox{Method}& Bunny & Genus6 & Rind & VSphere & Dino & VBunny
                             & Kangaroo & \\
    \midrule
    IDR - w/o Mask~\cite{yariv20} & \worst \na & \worst\na & \worst\na & \worst\na  & \worst\na & \worst\na & \worst\na  &\\
    $^\dagger$IDR - Phong~\cite{yariv20}                      & 9.67      & 3.71     & 10.24    & 16.93
                                      &3.71     & 14.80         & 2.77
                                      &\multirow{5}{*}{\minibox{\scriptsize
                                      Chamfer\\ \scriptsize $\times 10^{-3}$ \\
                              \scriptsize$\downarrow$ better}}\\
    $^\dagger$IDR - Neural~\cite{yariv20}                       & 9.84      & 1.35     &
    \best 0.21     &\best 0.16
                                      &2.07     & 9.11          & 5.43     &\\
	LSIG~\cite{nicolet21} & \best 0.06 & 2.85 & 3.94 & 4.78 & 2.09 & 4.66 & 1.80 &
    \\
	Ours & 0.18 &\best 0.12 & 5.56 & 3.71 &\best 1.25 &\best 0.10
         &\best 1.62 \\
	\midrule
    $^\dagger$IDR - Phong~\cite{yariv20}                      & 21.52 & 18.84  & 15.89 & 18.28         &
    20.01 & 17.27         & 21.67    &\multirow{5}{*}{\minibox{\scriptsize
    PSNR\\ \ \ \scriptsize dB\\ \scriptsize $\uparrow$ better}}\\
    $^\dagger$IDR - Neural~\cite{yariv20}                       & 23.10 & 28.70  & 26.24 & 25.70         &
    22.49 & 16.62         & 21.74    &\\
	LSIG~\cite{nicolet21} & \best 38.51 & 25.67 & 22.81 & 24.06 & 25.11 & 21.77 & 25.52 &  \\
	Ours & \best 38.86 & \best 32.94 & \best 28.46 & \best 30.50 & \best 28.16 &
    \best 29.62 & \best 26.48&\\ 
    \bottomrule
    \end{tabular}
}
    \caption{\label{tab:inverse} \textbf{Quantitative evaluation on inverse
        rendering of geometry.} An initial sphere is optimized to a diverse set of
        shapes from multi-view images. Chamfer distance is reported for
        geometric consistency and PSNR is reported to evaluate the visual
        appearance of optimized geometry. IDR~\cite{yariv20} w/o Mask does not
        converge for any of the shapes. 
        (marked \colorbox{red!10}{\na}). 
        Methods marked with $\dagger$ 
        require object masks.  LSIG~\cite{nicolet21} works well for a genus-0
        shape (\texttt{Bunny}) but struggles with high-genus shapes. 
        Best
        numbers are in \colorbox{orange!15}{orange}. 
        Shapes recovered using
        our method are shown on the top.
    }
    \vspace{-3em}
\end{table}

We propose an inverse-rendering method which uses a differentiable renderer
designed for triangle meshes to optimize geometry defined using parametric
level-sets.  As in the case of recent methods~\cite{li19cvpr,niemeyer20,yariv20}, we use
an analysis-by-synthesis approach. A photometric error comparing the captured
and the rendered images is minimized using gradient descent.  The gradients of
the error function are used to define an explicit flow-field. A corresponding
Eulerian deformation field is obtained to evolve the level-set function $\pphi$.
As a result we can take large steps in inverse rendering of
geometry with unconstrained topology and guarantees on mesh quality. The resulting
optimization is robust and does not require an object-mask as in the (unlike~\cite{niemeyer20,yariv20}).

We focus on geometry
recovery from synthetic scenes with known reflectance.
Our single-bounce forward rendering model uses a
collocated camera and point-light, with a known diffuse-Phong BRDF
Although we choose
Nvdiffrast~\cite{laine20} as the differentiable rasterizer in our method, in
theory it can be swapped with any other differentiable renderer.  Starting from
an initial estimate $\pphi$, for each time-step $t$ we first extract the
triangle mesh $\partial\Omega_L^t$ and minimize a photometric error $\mathcal{E}$.
We use the gradients of
$\mathcal{E}$ to define the flow-field as $\V^t(\x_i) = -\frac{\partial
\mathcal{E}}{\partial\x_i} - \lambda \textbf{L} \x_i,$ where $\textbf{L}\x_i$ is
used for smooth evolution. 
Taking a single descent step for $\straighttheta$ is sufficient since the
evolution does not need to follow a specific trajectory. 

\begin{wrapfigure}[8]{l}{0.35\textwidth} 
    \vspace{-1.5em}
    \begin{overpic}[width=0.35\textwidth, tics=5]{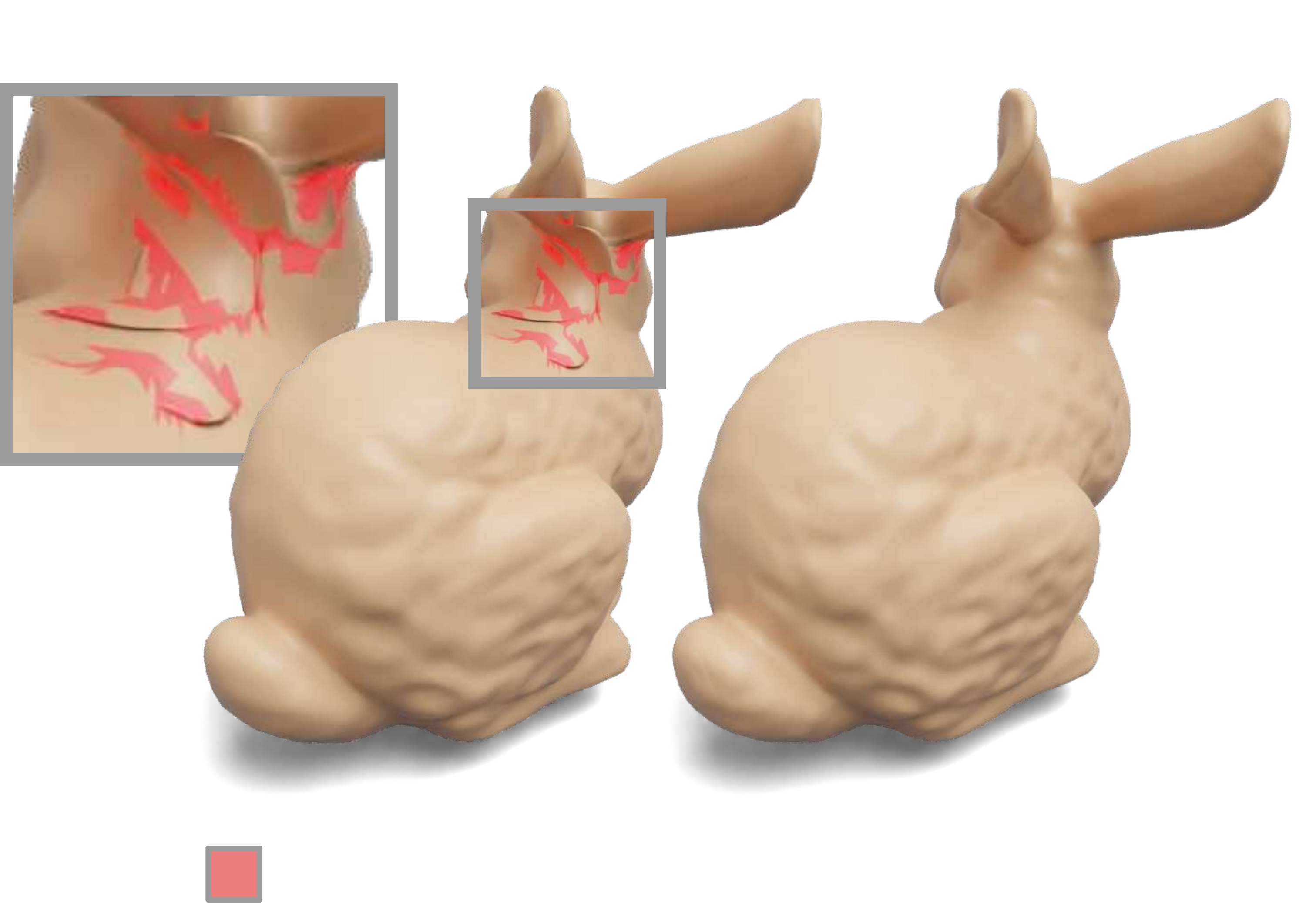}
        \put(45, 67) {\makebox[0pt]{\small LSIG~\cite{nicolet21}}}
        \put(78, 67) {\makebox[0pt]{\small Ours}}
        \put(23, 0) {\scriptsize Self Intersections}
    \end{overpic}
\end{wrapfigure}

\begin{figure}[ht]
    \centering
    \begin{overpic}[width=\linewidth, tics=5]{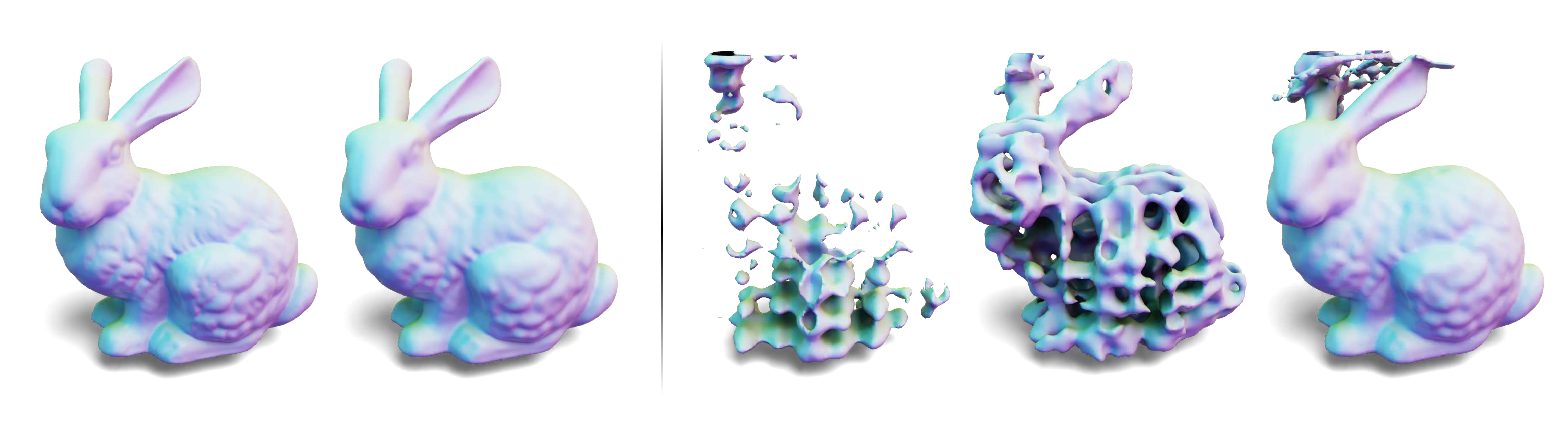}
        \put(70, 26) {\makebox[0pt]{IDR~\cite{yariv20}}}
        \put(30, 26) {\makebox[0pt]{Ours}}
        \put(10, 26) {\makebox[0pt]{Target}}
        \put(52, 22){\numcircle{1}}
        \put(73, 22){\numcircle{2}}
        \put(93, 22){\numcircle{3}}

        \put(11, 1){\makebox[0pt]{\scriptsize \textit{Mask}}}
        \put(11, -1){\makebox[0pt]{\scriptsize \textit{Shading}}}

        \put(52, 1) {\makebox[0pt]{\scriptsize No}}
        \put(71, 1) {\makebox[0pt]{\scriptsize Yes}}
        \put(90, 1) {\makebox[0pt]{\scriptsize Yes}}
        \put(30, 1) {\makebox[0pt]{\scriptsize No}}

        \put(52, -1) {\makebox[0pt]{\scriptsize Phong}}
        \put(71, -1) {\makebox[0pt]{\scriptsize Phong}}
        \put(90, -1) {\makebox[0pt]{\scriptsize Neural}}
        \put(30, -1) {\makebox[0pt]{\scriptsize Phong}}
    \end{overpic}
    \caption{\label{fig:idr}\textbf{Qualitative comparison for inverse rendering
        using implicit representations.} We evaluate IDR in three
        different settings. \numcircle{1} Without object-mask supervision it
        fails to converge to a reasonable geometry. \numcircle{2} With a
        known-reflectance model (Phong) the silhouette of the object is recovered but
    without any details. \numcircle{3} It requires a rendering network (unknown
reflectance) and an object-mask for good convergence---both of which not required for our method. 
}
\end{figure}

We evaluate the recovery on a diverse set of shapes, each of which is rendered from
$100$ random viewpoints. We use IDR~\cite{yariv20} and LSIG~\cite{nicolet21} as
 baselines. We test IDR in three settings, 1) w/o Mask, 2) With known Phong
shading, and 3) Using the Neural renderer in \cite{yariv20}.
Quantitative comparisons are in Table~\ref{tab:inverse} and qualitative
comparisons are in Fig.~\ref{fig:inverse_main}.
For a genus-0 shape (\texttt{Bunny}) LSIG~\cite{nicolet21} is able
to recover accurate geometry, but the optimized meshes can
have self intersections as shown on the left. It struggles with high-genus shapes as the
mesh connectivity remains static throughout the optimization routine. 
Even with correct topology at
initialization, the recovery is erroneous.
Comparisons with IDR~\cite{yariv20} are in
Figure~\ref{fig:idr}.

\subsection{User-defined Shape Editing}
\label{sec:user-editing}

\begin{wrapfigure}[8]{l}{0.38\textwidth} 
    \vspace{-1em}
    \begin{overpic}[width=0.38\textwidth, tics=5]{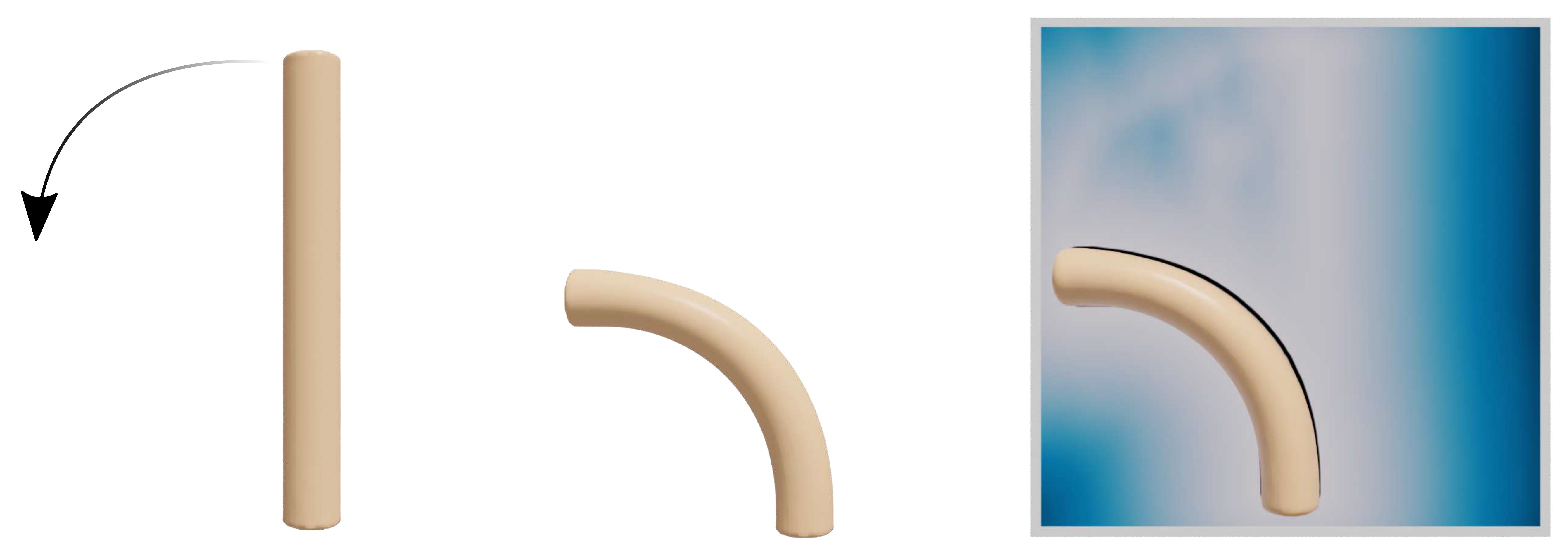}
        \put(40, 41) {\makebox[0pt]{\scriptsize GT (Explicit)}}
        \put(80, 41) {\makebox[0pt]{\scriptsize Ours (Implicit)}}
    \end{overpic}
    \vspace{-1.3em}
    \caption{\label{fig:deform}User-defined editing on parametric
    level-sets.}
\end{wrapfigure}

We demonstrate deformation operations on parametric level-sets
using constraints defined by a user. The problem setup is in line with the
extensive
literature~\cite{botsch2006deformation,sorkine2007rigid,sorkine2004laplacian,zayer2005harmonic}
on shape editing for triangle meshes. At $t=0$, we first extract a
mesh from a neural implicit surface using MC~\cite{lorensen87}. A user can specify
handle regions on this surface to either rotate, translate or freeze parts of
the shape along with their target locations. 
This generates a sparse deformation flow-field on the surface which we
\textit{densify} by minimizing a thin-shell energy function that penalizes local
stretching and bending on the surface~\cite{terzopoulos1987elastically}. More details are in the Appendix.
Estimating the flow-field requires solving a linear system $-k_S\Delta\V +
k_B\Delta^2\V = 0$, where $\Delta$ and $\Delta^2$ are the laplacian and
bi-laplacian operators.
$k_S$ and $k_B$ are weighting terms for stretching and
bending respectively.  
Additional constraints which adhere to user
specifications are also added to the linear system~\cite{botsch2007linear}. With
the obtained flow-field, we update the parameters of the level-set function such
that the surface propagation is as intended. We also use gradient regularization
as in~\cite{gropp20}. An example deformation is shown in
Fig.~\ref{fig:deform}.
\section{Discussion}
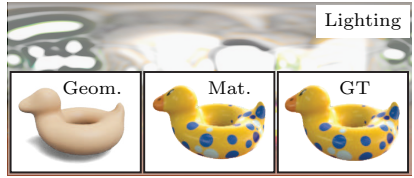
\begin{wrapfigure}[9]{l}{0.44\textwidth} 
    \vspace{-2em}
    \begin{overpic}[width=0.44\textwidth, tics=5]{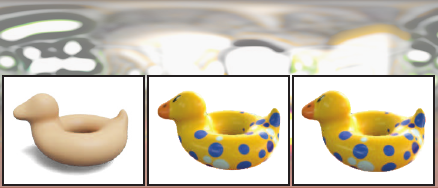}
        \put(21, 20) {\makebox[0pt]{\scriptsize Geom.}}
        \put(55, 20) {\makebox[0pt]{\scriptsize Mat.}}
        \put(86, 20) {\makebox[0pt]{\scriptsize GT}}
        \put(88, 37) {\makebox[0pt]{\colorbox{white}{\scriptsize Lighting}}}
    \end{overpic}
    \caption{\label{fig:joint_recovery} Joint recovery of geometry, complex material and lighting.}
\end{wrapfigure}

Our work formulates a level-set evolution method for parametrically defined
implicit surfaces. 
It does not require surface extraction to be differentiable and can be used to apply mesh algorithms
to neural implicit surfaces. We expect the proposed method to be particularly useful for inverse
problems. We showcase one example of joint recovery of geometry, material and lighting from
multi-view images in Fig.~\ref{fig:joint_recovery}, where we use our surface evolution method along
with components from~\cite{munkberg2021}. Although the surface deformation is as
dictated by the flow field, the corresponding implicit function may not retain gradient
characteristics during evolution. This could become a pertinent
problem for algorithms like sphere tracing~\cite{hart96} which require reliable distance queries,
and is an interesting avenue for future research. We hope this work encourages further inquiry into
recent work on geometry optimization by drawing connections to methods in computer vision and graphics
developed in the pre-deep-learning era.

\paragraph{\textbf{Acknowledgements}} This work was supported in part by NSF CAREER 1751365, NSF IIS 2110409, ONR
grant N000142012529, NSF Chase-CI grant 1730158, Adobe, Google, an Amazon Research Award, the Ronald L. Graham Chair and UC San Diego Center for Visual Computing. We thank Ceh Jan and
3dmixers users \texttt{roman\_hegglin} and \texttt{PhormaSolutions} for the 3D
models. We thank the anonymous reviewers for helpful comments and discussions.
\section{Appendix}
We organize the Appendix according to the section numbers used in the main
manuscript. In Section~\ref{sec:dsr}, the proofs for Result 3 and Result 4 are
outlined. In Section~\ref{sec:curvature}, implementation details are added
along with additional results in
\hyperref[fig:smooth_progress]{\greendashuline{Figure~\ref{fig:smooth_progress}}}
and \hyperref[fig:smooth_all]{\greendashuline{Figure~\ref{fig:smooth_all}}}. In
Section~\ref{sec:inverse_rendering}, implementation details for our inverse
rendering experiments are discussed and qualitative comparisons with previous
works are shown in
\hyperref[fig:inverse_all]{\greendashuline{Figure~\ref{fig:inverse_all}}}.
An inverse rendering experiment using LSIG~\cite{nicolet21} with different
geometry initializations is shown in
\hyperref[fig:topology_lsig]{\greendashuline{Figure~\ref{fig:topology_lsig}}}.
Section~\ref{sec:editing} describes the user-defined shape editing method in
more detail. A comparison with NFGP~\cite{yang21} is in 
\hyperref[fig:deform_evolution]{\greendashuline{Figure~\ref{fig:deform_evolution}}}.
\setcounter{section}{5}
\setcounter{subsection}{1}
\subsection{Differentiable Surface Rendering}
\label{sec:dsr}
We first draw comparisons between Differentiable Volumetric
Rendering\footnote{Here,
volumetric rendering is a misnomer. It's really surface rendering for geometry
defined with volume/occupancy.}~\cite{niemeyer20}, Implicit Differentiable
Renderer (IDR)~\cite{yariv20} and Differentiable Iso-Surface Extraction in Result
3. Next, in Result 4, we show that methods in~\cite{niemeyer20,yariv20} deviate
from the level-set theory for tangential flow fields.
\setcounter{result}{2}
\begin{result}
Surface evolution using differentiable ray-marching of parametric implicit
surfaces~\cite{mescheder19,yariv20} is the same as using differentiable
iso-surface extraction~\cite{remelli20} when the viewing direction
\textnormal{$\textbf{v}_u$} is
parallel to the normal \textnormal{$\n$} at the intersection point \textnormal{$\x_u$}. The parameters $\straighttheta$
for the level-set function $\pphi$ are updated as:
\end{result}
$$
    \straighttheta \leftarrow \straighttheta - \lambda \sum_{\x_u} \V\cdot
    \nabla\pphi\frac{\partial\pphi}{\partial\straighttheta}.
$$
\begin{proof}
Inverse rendering methods in~\cite{niemeyer20,yariv20} find a surface-intersection
point by marching a along ray that is spawned from the camera centre
$\textbf{c}$, in the direction $\textbf{v}_u$, for each pixel $u$. The
intersection point $\x_u$ is analytically defined as:
\begin{align}
\x_u = \textbf{c} + d_u\textbf{v}_u,
\label{supp_eq:ray-march}
\end{align}
where $d_u$ is the depth for $\x_u$.
The distance $d_u$ in~\cite{niemeyer20} is estimated from the camera centre. For
IDR~\cite{yariv20}, the distance is computed from a
point $\textbf{y}$ close to the surface which modifies (\ref{supp_eq:ray-march}) to $\x_u =
\textbf{y}_u + d_u\textbf{v}_u$ ($\textbf{y}_u$ is denoted as $\x_0$ in
\cite{yariv20}). In both the methods, a rendering loss function $\mathcal{L}$
(photometric error) is
computed for pixels $u \in U$. To update the geometry parameters
$\straighttheta$, the gradient
$\frac{\partial\mathcal{L}}{\partial\straighttheta}$ is computed:
\begin{align}
\frac{\partial\mathcal{L}}{\partial\straighttheta} =
\sum_{u \in U}
\frac{\partial\mathcal{L}}{\partial\x_u}\frac{\partial\x_u}{\partial\straighttheta}
= \sum_{u \in U}
\frac{\partial\mathcal{L}}{\partial\x_u}\frac{\partial\x_u}{\partial
d_u}\frac{\partial d_u}{\partial\straighttheta}
= \sum_{u \in U}\frac{\partial\mathcal{L}}{\partial\x_u}\cdot \textbf{v}_u\frac{\partial
d_u}{\partial\straighttheta}
\label{supp_eq:loss}
\end{align}
To compute $\frac{\partial d_u}{\partial\straighttheta}$, we slightly deviate from the
derivations in~\cite{niemeyer20,yariv20} for clarity. For points $\x_u$ on the implicit
surface, we know $\pphi(\x_u) = 0$. Using implicit differentiation:
\begin{align}
&\qquad \frac{\partial\pphi}{\partial\straighttheta} +
\frac{\partial\pphi}{\partial\x_u}\frac{\partial\x_u}{\partial\straighttheta}
= 0 \nonumber\\
&\iff \frac{\partial\pphi}{\partial\straighttheta} +
\frac{\partial\pphi}{\partial\x_u} \frac{\partial\x_u}{\partial d_u}\frac{\partial d_u}{\partial\straighttheta}
= 0 \nonumber\\
&\iff \frac{\partial\pphi}{\partial\straighttheta} +
\frac{\partial\pphi}{\partial\x_u}\cdot \textbf{v}_u \frac{\partial d_u}{\partial\straighttheta}
= 0\nonumber \qquad\lhd\ \text{From (\ref{supp_eq:ray-march})}\\
&\iff \frac{\partial d_u}{\partial\straighttheta} =
-\frac{1}{\frac{\partial\pphi}{\partial\x_u}\cdot\textbf{v}_u
}\frac{\partial\pphi}{\partial\straighttheta}
\label{supp_eq:implicit}
\end{align}
From (\ref{supp_eq:loss}) and (\ref{supp_eq:implicit}) we have:
\begin{align}
\frac{\partial\mathcal{L}}{\partial\straighttheta} =
 - \sum_{u \in U}\frac{\partial\mathcal{L}}{\partial\x_u}\cdot
\frac{\textbf{v}_u}{\frac{\partial\pphi}{\partial\x_u}\cdot\textbf{v}_u}\frac{\partial\pphi}{\partial\straighttheta}
 \approx - \sum_{u \in U}\frac{\partial\mathcal{L}}{\partial\x_u}\cdot
\frac{\nabla\pphi}{\left| \nabla\pphi
\right| ^2}\frac{\partial\pphi}{\partial\straighttheta} \ (\text{when}\ \textbf{v}_u
\to \nabla\pphi)
\label{supp_eq:parallel}
\end{align}
Taking $-\frac{\partial\mathcal{L}}{\partial\x_u} = \V$, we can update the
parameters $\straighttheta$ using (\ref{supp_eq:parallel}) and gradient descent as:
\begin{align}
    \straighttheta \leftarrow \straighttheta - \lambda \sum_{\x_u} \V\cdot
    \nabla\pphi\frac{\partial\pphi}{\partial\straighttheta}.
    \label{supp_eq:dvr_meshsdf}
\end{align}
Equation (\ref{supp_eq:dvr_meshsdf}) shows that DVR/IDR and MeshSDF$^\straighttheta$
are closely related, while Result 1 shows that these methods do not agree with the
level-set theory. $\square$
\end{proof}
\begin{result}
    \label{result:dvr_lse}
    Differentiable ray-marching of parametric implicit
    surfaces~\cite{mescheder19,yariv20} disagrees with the level-set equation for
    tangential components \textnormal{$\V^\perp$} of the flow field
    \textnormal{$\V$}. The change in parameters
    $\Delta\straighttheta$ is:
\end{result}
\begin{align}
   \Delta\straighttheta = \lambda\sum_{\x_u}\pm|  \V^\perp
   | \tan (\arccos (\nabla\pphi \cdot
   \textbf{v}_u))\frac{\partial\pphi}{\partial\straighttheta}.
   \label{supp_eq:dvr_tangent}
\end{align}
\begin{proof}
    From (\ref{supp_eq:parallel}), we know the change in parameters $\straighttheta$
    for methods in~\cite{niemeyer20,yariv20} is:
    \begin{align}
        \Delta\straighttheta = \lambda\sum_{\x_u}
        \V\cdot
        \frac{\textbf{v}_u}{\nabla\pphi\cdot\textbf{v}_u}
        \frac{\partial\pphi}{\partial\straighttheta}.
    \end{align}
    When the flow-field is tangential to the surface, $\V = \V^{\perp}$, and
    $\V^{\perp} \perp \nabla\pphi$. The change in parameters for this flow-field
    is:
    \begin{align}
        \Delta\straighttheta = \lambda\sum_{\x_u}
        \frac{\V^{\perp}\cdot\textbf{v}_u}{\nabla\pphi\cdot\textbf{v}_u}
        \frac{\partial\pphi}{\partial\straighttheta}.
        \label{supp_eq:perp-1}
    \end{align}
    When $\pphi:\mathbb{R}^2 \mapsto \mathbb{R}$ is a level-set function defined
    in 2D, we can modify (\ref{supp_eq:perp-1}) as:
    \begin{align}
        \Delta\straighttheta &= \lambda\sum_{\x_u}
        \pm|\V^{\perp}|\frac{\sin\alpha}{\cos\alpha}
        \frac{\partial\pphi}{\partial\straighttheta}\qquad\ \text{where, $\alpha =
        \arccos(\nabla\phi\cdot\textbf{v}_u)$ and $\pphi$ is an SDF} \nonumber\\
                             &= \lambda\sum_{\x_u}
        \pm|\V^{\perp}|\tan\alpha\
        \frac{\partial\pphi}{\partial\straighttheta}\nonumber\\
                             &= \lambda\sum_{\x_u}
        \pm|\V^{\perp}|\tan(\arccos(\nabla\pphi\cdot\textbf{v}_u))\
        \frac{\partial\pphi}{\partial\straighttheta}.
        \label{supp_eq:perp-2}
    \end{align}
    The parameters here could change depending on the angle subtended between
    the normal $\nabla\pphi$ and the viewing direction $\textbf{v}_u$. This
    dependence on the viewing direction is a result of using differentiable
    ray-marching. When the function is deformed using the level-set method
    described in the main paper, the change in parameters $\Delta\straighttheta$
    does not depend on the viewing direction and is $0$. For 3D level-set
    functions, the parameters could still change as the term
    $\V^{\perp}\cdot\textbf{v}_u$ in (\ref{supp_eq:perp-1}) could be non-zero.
\end{proof}
\setcounter{section}{6}
\setcounter{subsection}{0}
\subsection{Curvature-based Deformation}
\label{sec:curvature}
\begin{figure}[t]
    \vspace{1em}
    \centering
    \begin{overpic}[width=\linewidth, tics=5]{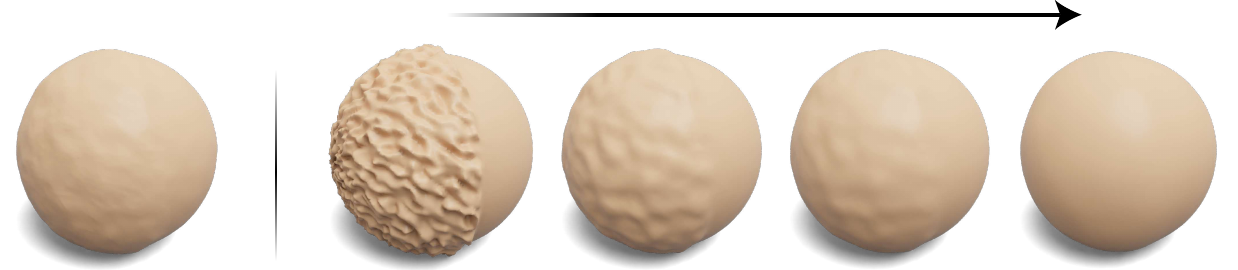}
        \put(60, 22){\small Ours}
        \put(4, 22){\small NFGP~\cite{yang21}}
    \end{overpic}
    \caption{\label{fig:smooth_progress}\textbf{Surface smoothing evolution on a
    half-noisy sphere.} (\textit{Right}) Our method can evolve a noisy surface using an explicit
diffusion flow-field. This allows smoothing on implicitly defined surfaces with
increasing levels of smoothness. (\textit{Left}) Yang~\etal's~\cite{yang21} method uses an
optimization objective for a single level of smoothness.}
\end{figure}

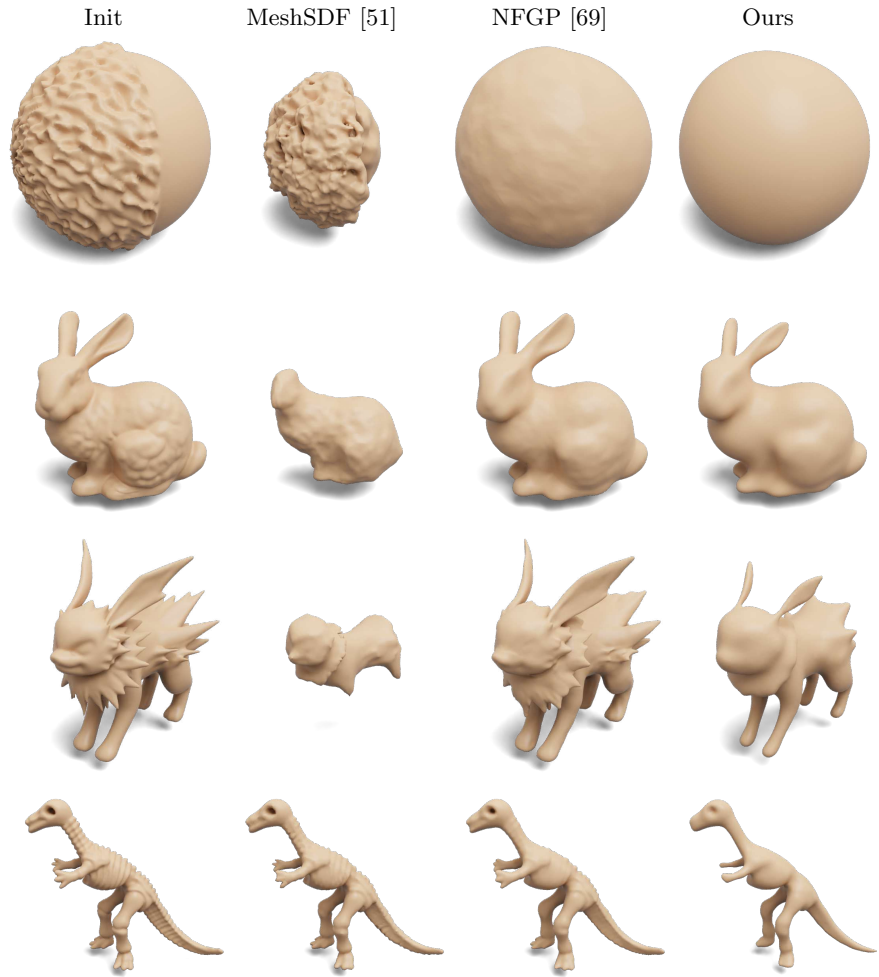
\begin{figure}[H]
    \centering
    \begin{overpic}[width=0.98\linewidth, tics=5]{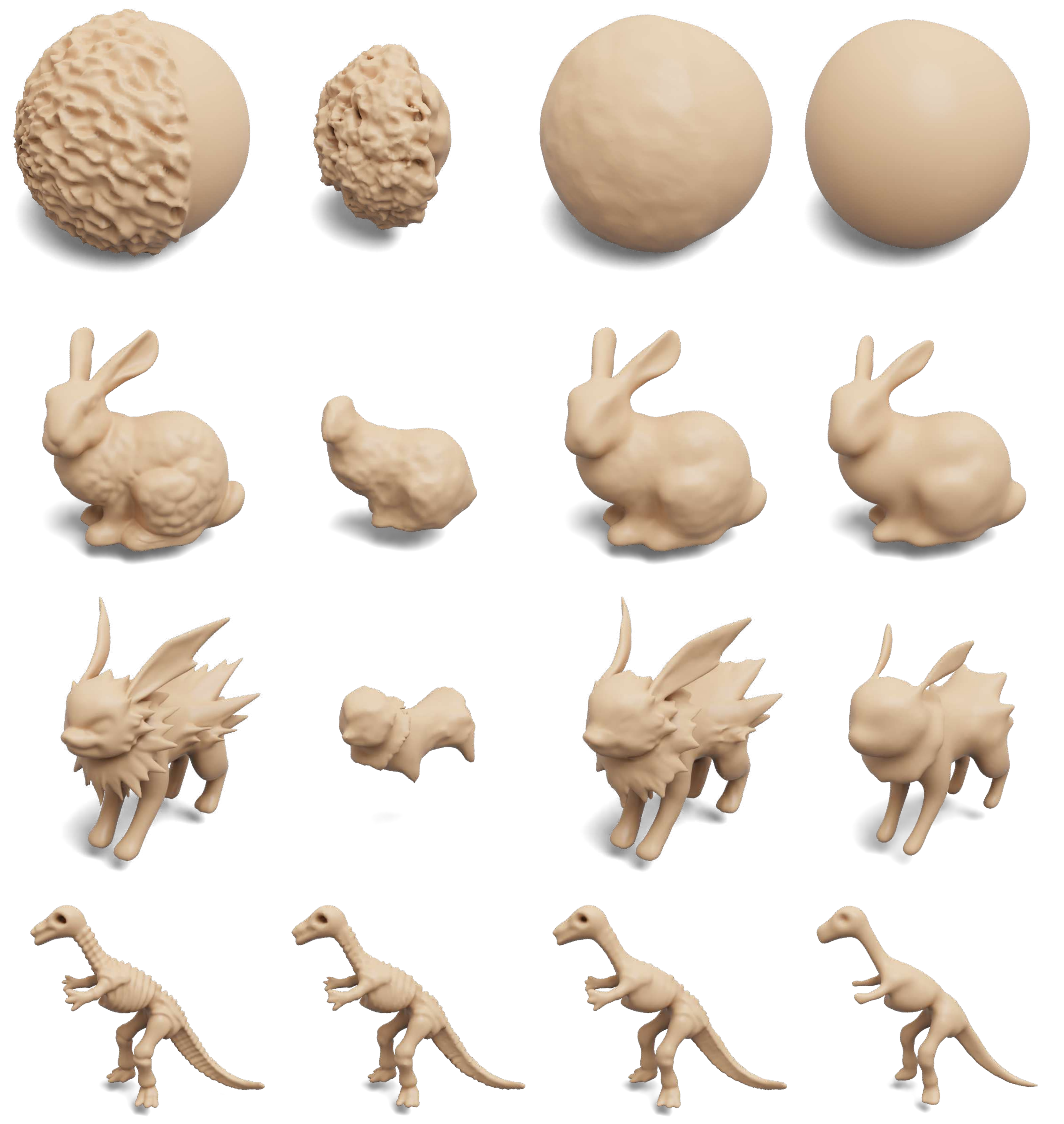}
        \put(11, 101) {\makebox[0pt]{Init}}
        \put(34, 101) {\makebox[0pt]{MeshSDF~\cite{remelli20}}}
        \put(58, 101) {\makebox[0pt]{NFGP~\cite{yang21}}}
        \put(81, 101) {\makebox[0pt]{Ours}}
    \end{overpic}
    \caption{\label{fig:smooth_all}\textbf{Qualitative comparisons for surface
    smoothing.} We apply surface smoothing on four computer graphics models
encoded as parametric level-sets. Best viewed when zoomed-in.}
\end{figure}

For each shape, we optimize a SIREN~\cite{sitzmann19} MLP with 5 layers, each of
which has a 512-sized vector output.  We use the  publicly-released
code\footnote{https://github.com/stevenygd/NFGP} by Yang~\etal~\cite{yang21} for
SDF queries and optimization of the network. The flow-field defined in Equation
(14) is used for smoothing. The learning rate is $10^{-6}$, the time-delta
$\Delta t$ is $0.95$ and an eikonal regularization term (enforcing $\nabla\pphi$
= 1)~\cite{gropp2020implicit} is used with the error function defined in Equation (5). The regularization
term is weighed $10^{-3}$ times against the main objective. We use Adam
optimizer~\cite{kingma2014adam} and use 200 gradient steps for each time step.
The Lagrangian surface is extracted using Marching Cubes at $(120 \pm 3)^3$
resolution. Figure~\ref{fig:smooth_progress} shows
the surface evolution with respect to time on a half-noisy sphere (similar
to~\cite{taubin1995signal}). In addition to the results in the main manuscript,
we show qualitative comparisons with other methods in
Figure~\ref{fig:smooth_all}.

\clearpage
\subsection{Inverse Rendering}
\label{sec:inverse_rendering}
\begin{wrapfigure}[12]{r}[-4pt]{0.4\textwidth}
    \vspace{-1em}
    \begin{overpic}[width=0.35\textwidth, tics=15]{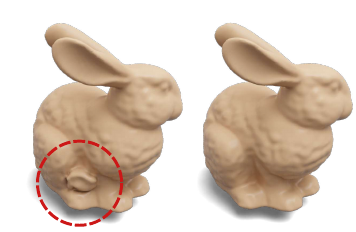}
        \put(52, 73){\makebox[0pt]{Smoothness Term}}
        \put(28, 62){\makebox[0pt]{\small without}}
        \put(75, 62){\makebox[0pt]{\small with}}
    \end{overpic}
    \caption{\label{fig:ablation}Ablation on smoothness regularization term in
    the flow-field for inverse rendering.}
\end{wrapfigure}

\begin{figure}[t]
    \centering
    \begin{overpic}[width=0.6\textwidth, tics=5]{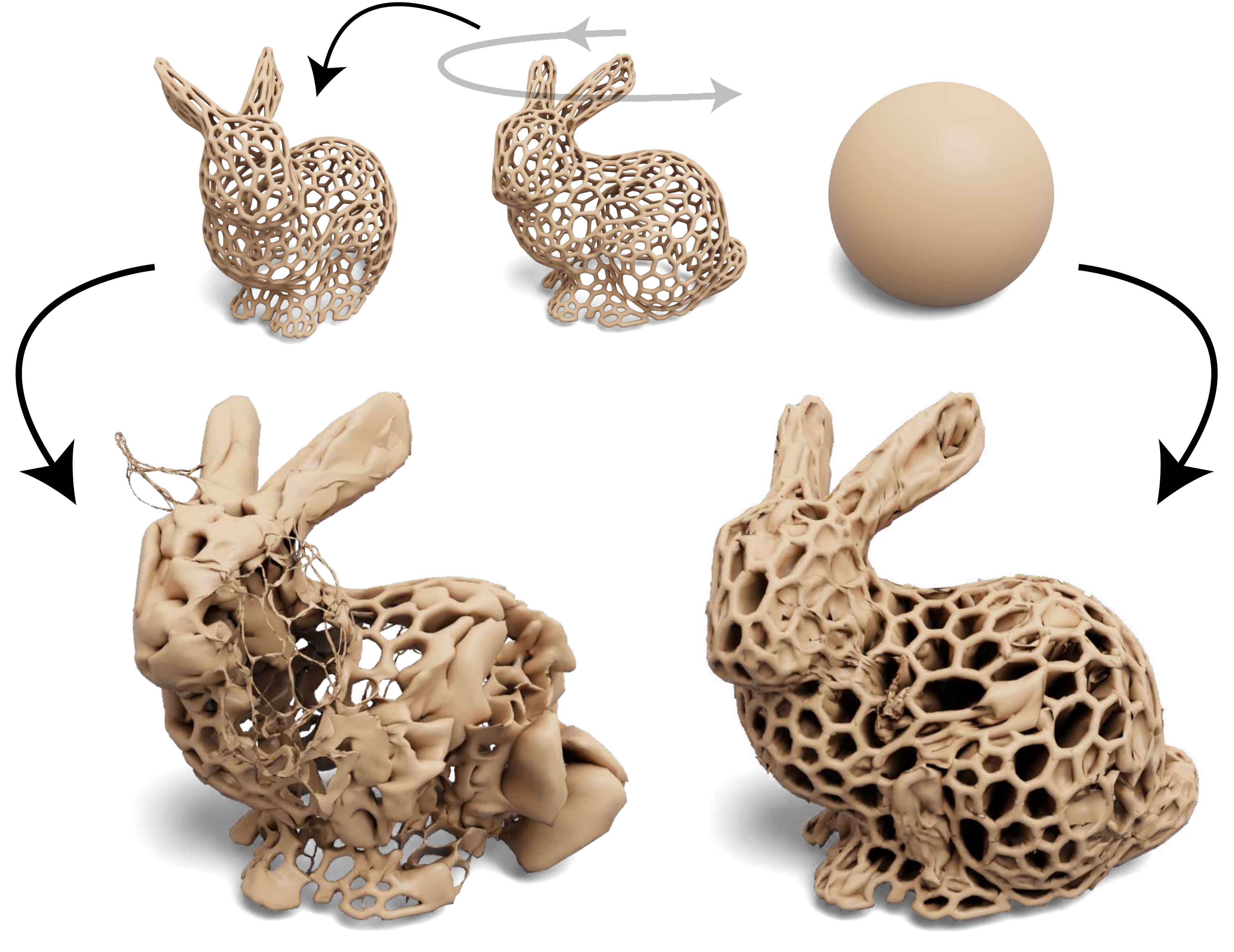}
        \put(20, 76) {\makebox[0pt]{Init 1}}
        \put(50, 76) {\makebox[0pt]{GT}}
        \put(59, 72) {\makebox[0pt]{$45^{\text{o}}$}}
        \put(76, 76) {\makebox[0pt]{Init 2}}
        \put(5, 65) {\makebox[0pt]{\small Correct}}
        \put(5, 60) {\makebox[0pt]{\small Topology}}
        \put(97, 65) {\makebox[0pt]{\small Different}}
        \put(97, 60) {\makebox[0pt]{\small Topology}}
    \end{overpic}
    \caption{\label{fig:topology_lsig}\textbf{LSIG~\cite{nicolet21} converges to
    a poor estimate with correct topology at initialization.} Optimizing a
triangle mesh for inverse rendering problems is non-trivial even when the
initialization is close to the ground truth and has correct topology. For
\texttt{VBunny} we rotate the target geometry and use it as initialization.
The shape optimized using LSIG~\cite{nicolet21} has correct topology but with poor mesh quality and
reconstruction accuracy.}
\end{figure}

\begin{figure}[ht!]
    \centering
    \vspace{1em}
    \begin{overpic}[width=\linewidth, tics=5]{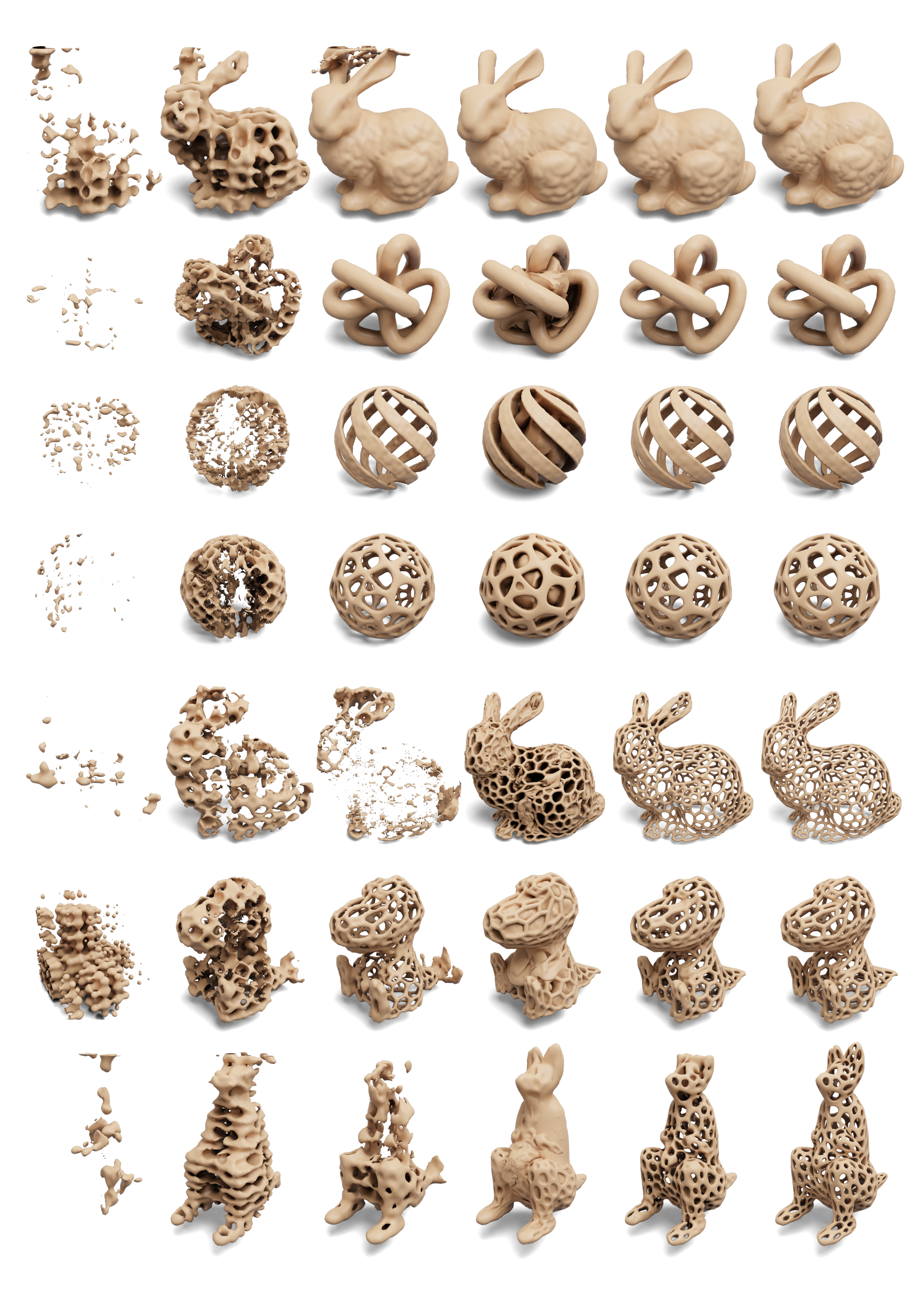}
        \put(5, 102) {\makebox[0pt]{IDR}}
        \put(17, 102) {\makebox[0pt]{IDR}}
        \put(29, 102) {\makebox[0pt]{IDR}}
        \put(41, 102) {\makebox[0pt]{LSIG}}
        \put(52, 102) {\makebox[0pt]{Ours}}
        \put(64, 102) {\makebox[0pt]{GT}}

        \put(5, 100) {\makebox[0pt]{\textcolor{OliveGreen}{No}}}
        \put(17, 100) {\makebox[0pt]{\textcolor{red}{Yes}}}
        \put(29, 100) {\makebox[0pt]{\textcolor{red}{Yes}}}
        \put(41, 100) {\makebox[0pt]{\textcolor{OliveGreen}{No}}}
        \put(52, 100) {\makebox[0pt]{\textcolor{OliveGreen}{No}}}
        \put(64, 100) {\makebox[0pt]{\textit{Mask}}}

        \put(5, 98) {\makebox[0pt]{Neural}}
        \put(17, 98) {\makebox[0pt]{Phong}}
        \put(29, 98) {\makebox[0pt]{Neural}}
        \put(41, 98) {\makebox[0pt]{Phong}}
        \put(52, 98) {\makebox[0pt]{Phong}}
        \put(64, 98) {\makebox[0pt]{\textit{Shading}}}
    \end{overpic}
    \caption{\label{fig:inverse_all}\textbf{Qualitative comparisons for Inverse Rendering of Geometry.}}
\end{figure}
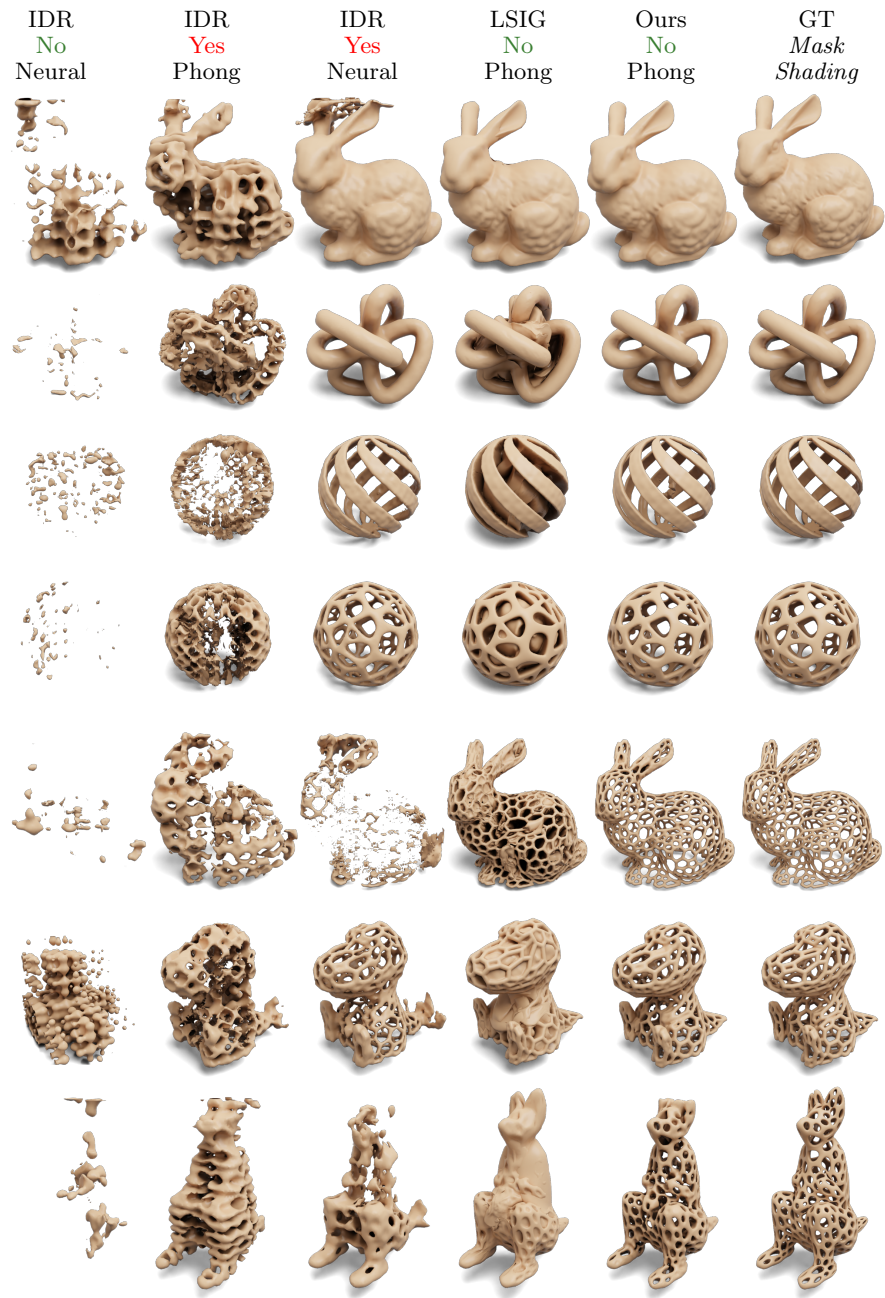

Our method uses the same spherical initialization for all the shapes. The
network is composed of 4 layers, with 512 neurons in each layer. The
learning-rate is $2\times 10^{-6}$, the weight-decay factor is $0.1$ and the
time-delta $\Delta t$ is $10^{-4}$. Each object is confined in a $2^3$ volume
and is rendered with Nvdiffrast~\cite{laine20}. The weight for the smoothness
regularization term linearly decreases from $10^{-3}$ to $0$. An ablation is
shown in Figure~\ref{fig:ablation}. We do not observe reliable improvements
quantitatively with the regularizer, but do observer better convergence. The
Phong shading model is used with $0.55$ as the albedo value. We show an
experiment with different topological initializations for LSIG~\cite{nicolet21}
in Figure~\ref{fig:topology_lsig}. All the qualitative comparisons for results
in Table 1 (main) are in Figure~\ref{fig:inverse_all}.

\subsection{User-defined Deformation}
As discussed in the main paper, we use the thin-shell
energy~\cite{terzopoulos1987elastically} loss to densify a
sparsely defined flow-field from user inputs. It induces a flow field
which is characterized using an Euler-Lagrange equation: 
\begin{align}
    -k_s\Delta \V + k_b \Delta^2 \V = 0.
    \label{supp_eq:eul_lag_v}
\end{align}
We know $\V$ for the regions on the surface where the user specifies the
deformation and would like to estimate it for the entire surface such that there
is minimum bending and stretching. 
Equation~\ref{supp_eq:eul_lag_v}, however is a result of a linearized version of the
thin-shell energy loss. In this case, the resultant deformations can be
unpleasant.  This is required for explicit surface deformation methods as they
are expected to be conducive to real-time editing. On the contrary, our goal is
to have plausible deformations which can be slightly more time consuming. For
each editing operation ($\x \to \x'$), we break the deformation in $T$ time
steps as $\x \to \x^1 \to \x^2 \to \x^3 \dots \x^T$. For each time-step $t$, we
solve for (\ref{supp_eq:eul_lag_v}) (using a sparse linear solver) with the
following user
constraints:
\begin{align}
\V(\x_h^{t-1}) =
\x_h^{t} - \x_h^{t-1},
\label{supp_eq:v_t}
\end{align}
where $\x_h \in \mathcal{H}$ belongs to a set of handle vertices for which the
user defines deformation. Having obtained the flow-field for all the points on
the surface, we evolve the surface using an Euler step as $\x^t = \x^{t-1} +
\V(\x^{t-1})$. Note the absence of a delta term ($\Delta t$) in the Euler step.
This is because of how we define the flow-field; we know exactly where the
surface is expected to be at a given time step (From (\ref{supp_eq:v_t})). This is
different from gradient-based Euler integration where we take small steps in the
direction of the flow-field. Since we know \textit{exactly} where the surface is
at time $t$ (can assume $\pphi(\x) = 0$ for surface points), we can tweak the
objective defined in Equation 5 (main):
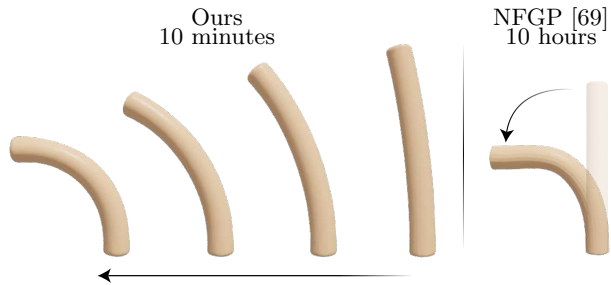
\begin{figure}[b!]
    \centering
    \vspace{2em}
    \begin{overpic}[width=0.7\linewidth, tics=5]{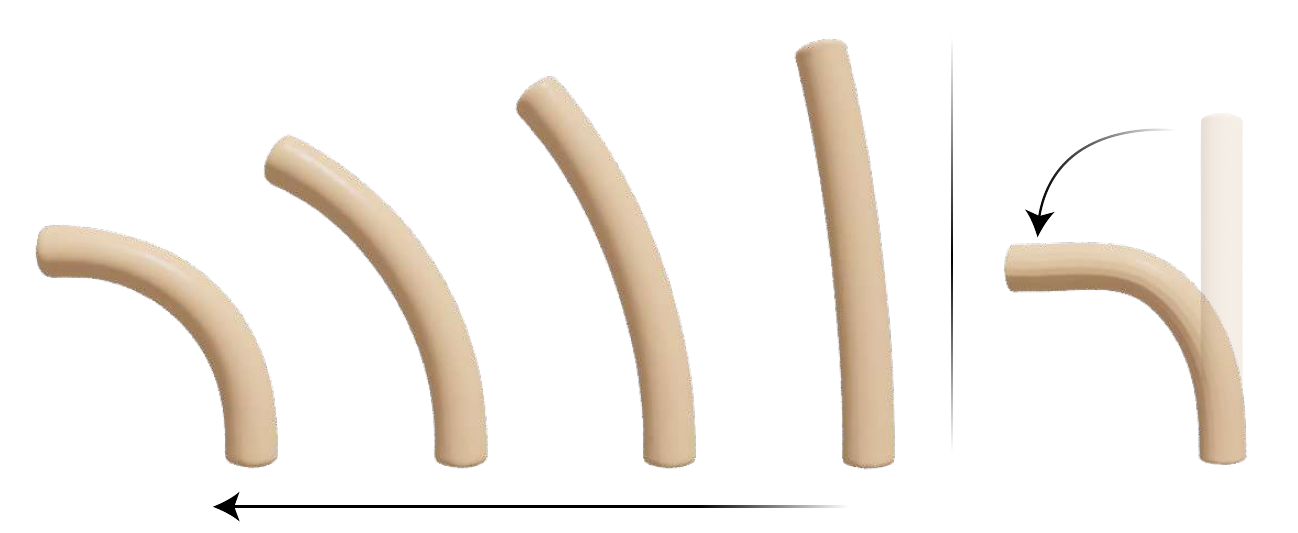}
        \put(87, 41) {\makebox[0pt]{NFGP~\cite{yang21}}}
        \put(35, 41) {\makebox[0pt]{Ours}}
        \put(87, 38) {\makebox[0pt]{10 hours}}
        \put(35, 38) {\makebox[0pt]{10 minutes}}
    \end{overpic}
    \caption{\label{fig:deform_evolution}\textbf{User-defined rotation on an implicitly
    defined cylinder.} (\textit{Left}) User editing using our method is faster
(takes 10 min) than (\textit{Right}) NFGP which takes 10 hr for the same editing
operation.}
\end{figure}

\begin{align}
    \min_\straighttheta J = \frac{1}{|\partial\Omega_L^t|}\sum_{\x\in\Omega_L^t}
    ||\pphi(\x)||^2 + \beta\sum_{\x\in\Omega} (|\nabla\pphi(\x)| - 1)^2,
    \label{supp_eq:new_J}
\end{align}
where we also add Eikonal regularization. An example deformation is shown in
Figure~\ref{fig:deform_evolution} including a comparison with
NFGP~\cite{yang21}. Note that our goal with these experiments is to showcase the
versatility of the level-set method and not propose a method which performs
accurate and real-time deformations; we would still pose this application of
user-editing as a proof of concept. Several explicit-surface based
methods~\cite{botsch10} exist
which can probably generate better deformations.
\paragraph{Implementation Details} We use pre-trained neural implicit
representations provided in the publicly-released code by
Yang~\etal~\cite{yang21}. The number of time steps $T$ is 20. The learning rate
is $2\times 10^{-6}$ and 750 gradient steps are taken to minimize
Equation~\ref{supp_eq:new_J} using Adam~\cite{kingma2014adam}. The regularization
weighting term $\beta = 10^{-4}$.
\label{sec:editing}
\begin{figure}[t]
\centering
    \begin{overpic}[width=0.38\textwidth, tics=5]{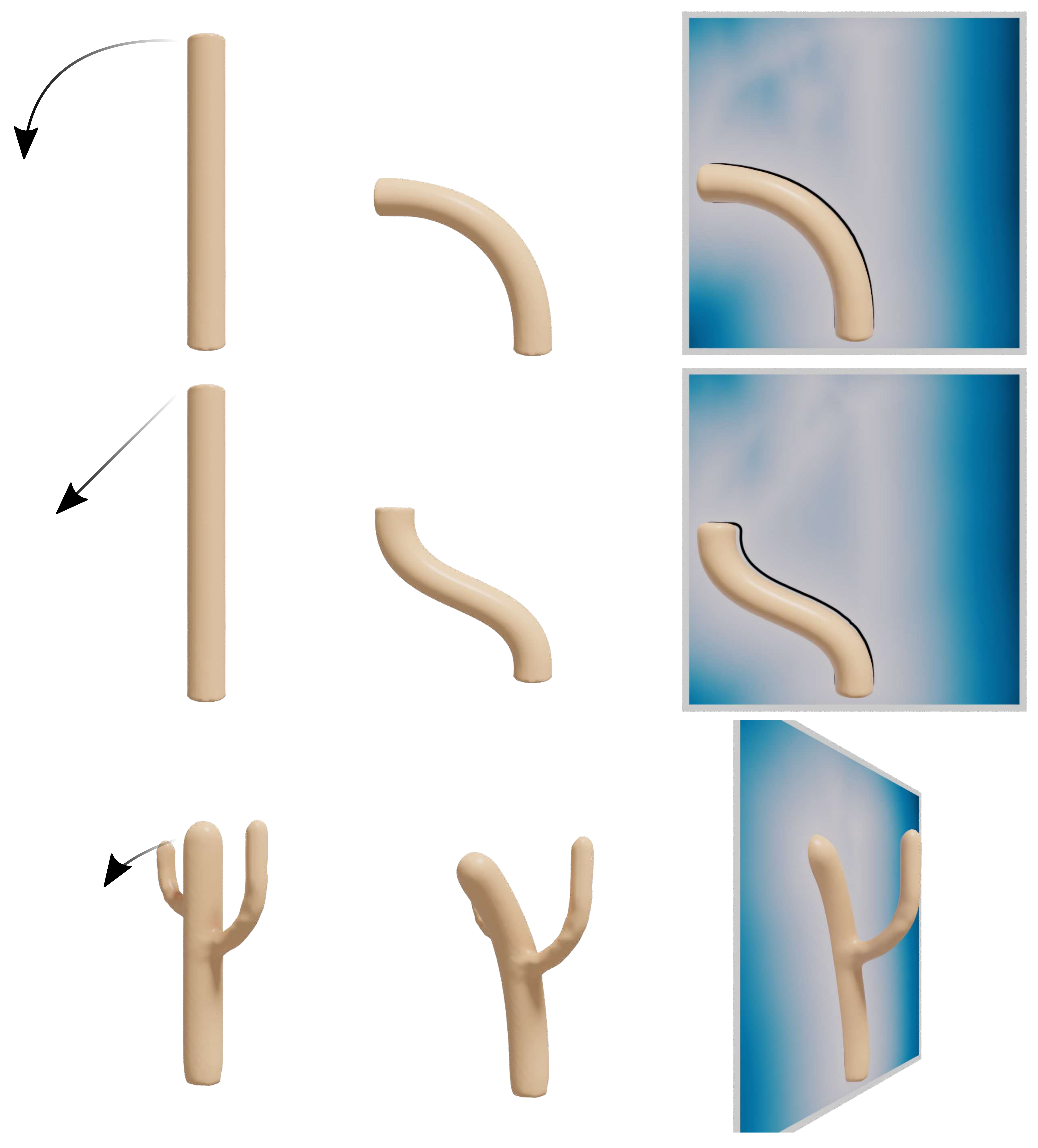}
        \put(40, 101) {\makebox[0pt]{\scriptsize GT (Explicit)}}
        \put(75, 101) {\makebox[0pt]{\scriptsize Ours (Implicit)}}
    \end{overpic}
    \vspace{-1.3em}
    \caption{\label{fig:deform}User-defined editing on parametric
    level-sets.}
    \end{figure}

\bibliographystyle{splncs04}
\bibliography{main}
\end{document}